\pdfoutput=1

\documentclass[11pt]{article}

\usepackage[final]{acl}

\usepackage{times}
\usepackage{latexsym}

\usepackage[T1]{fontenc}

\usepackage[utf8]{inputenc}

\usepackage{microtype}

\usepackage{inconsolata}

\usepackage{graphicx}
\usepackage{multirow}
\usepackage{makecell}
\usepackage{tabularx}
\usepackage{booktabs}
\newcommand{\tr}[1]{\textcolor{red}{#1}}

\usepackage{tcolorbox}
\usepackage{listings}
\usepackage{xcolor}
\usepackage{fancyvrb}
\definecolor{lightblue}{RGB}{46,110,187}
\definecolor{darkred}{RGB}{150,38,31}
\definecolor{darkgreen}{rgb}{0.0, 0.75, 0.0}
\definecolor{blue}{HTML}{3572EF}

\usepackage{lipsum}
\usepackage{longtable}
\usepackage{nicematrix}
\usepackage{pifont}
\usepackage{tikz}
\usepackage{enumitem}

\title{A Computational Approach to Visual Metonymy}

\author{Saptarshi Ghosh, Linfeng Liu, Tianyu Jiang \\
University of Cincinnati \\
\texttt{\{ghosh2si,liu2lf\}@mail.uc.edu, tianyu.jiang@uc.edu} \\
}

\begin{document}
\maketitle
\begin{abstract}

Images often communicate more than they literally depict: a set of tools can suggest an occupation and a cultural artifact can suggest a tradition. This kind of indirect visual reference, known as visual metonymy, invites viewers to recover a target concept via associated cues rather than explicit depiction. In this work, we present the first computational investigation of visual metonymy. We introduce a novel pipeline grounded in semiotic theory that leverages large language models and text-to-image models to generate metonymic visual representations. Using this framework, we construct ViMET,\footnote{\url{https://github.com/cincynlp/ViMET}} the first visual metonymy dataset comprising 2,000 multiple-choice questions to evaluate the cognitive reasoning abilities in multimodal language models. Experimental results on our dataset reveal a significant gap between human performance (86.9\%) and state-of-the-art vision-language models (65.9\%), highlighting limitations in machines' ability to interpret indirect visual references.

\end{abstract}

\section{Introduction}

\begin{figure}[ht]
    \centering
    \includegraphics[width=0.98\linewidth]{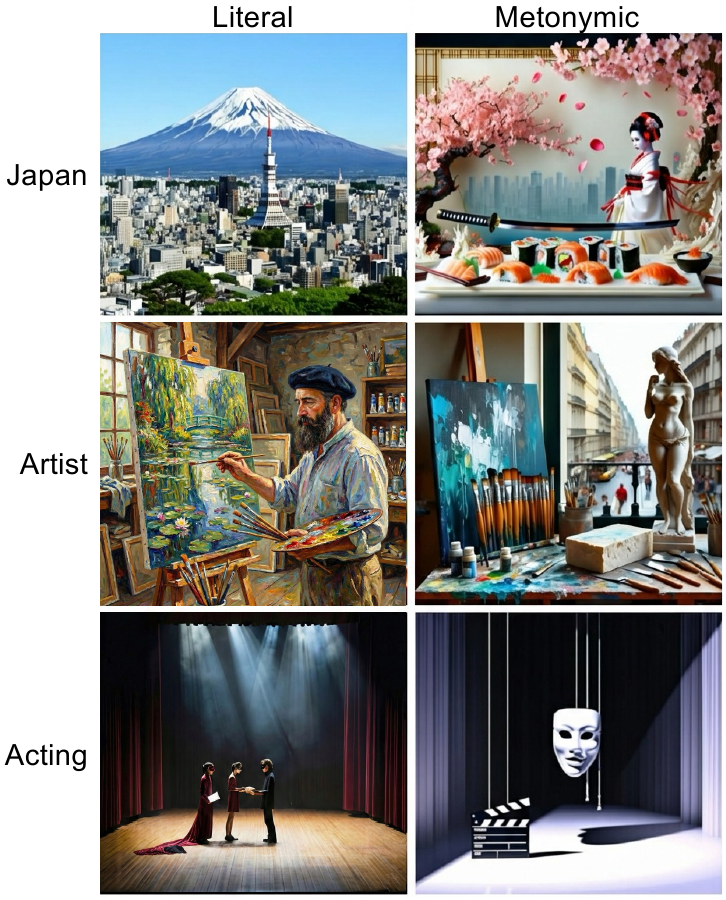}
    \caption{Examples of text-to-image model generated literal and metonymic images for the concept words \textit{Japan, Artist}, and \textit{Acting}. The literal image depicts the concept explicitly, while the metonymic image evokes the idea of the concept through cultural, contextual, or symbolic association.}
    \label{fig:intro_img}
\end{figure}

Visual metonymy is a form of indirect representation in which an image evokes a concept not by depicting it directly, but by presenting visually associated cues that invite the viewer to infer the intended meaning~\citep{visual_metonymy}. For example, an image of a canvas and paint brush evokes the idea of an artist without the artist being shown. Previous studies on linguistic metonymy have shown that meaning arises through associative links between a concept and a related cue within the same conceptual domain~\citep{lakoff1980, radden}. The principle holds in the visual modality as well, where the viewer infers a broader concept from a related visual cue through associative links. Unlike textual metonymy, which unfolds sequentially through language, visual metonymy communicates meaning instantaneously and often more ambiguously, relying heavily on the observer’s interpretive framework~\citep{forceville2009nonverbal}.

Figure \ref{fig:intro_img} illustrates visual metonymic images, comparing them with a literal depiction of the concept words \textit{Japan}, \textit{Artist}, and \textit{Acting}. The literal image depicts the concept through explicit visual representation, while the metonymic image conveys the concept indirectly by depicting associated objects. These associations rely on conceptual links---cultural, contextual, or symbolic---through which the viewer infers the intended idea without it being explicitly shown. The first image illustrates \textit{cultural association}, where the concept of \textit{Japan} is evoked using objects related to Japanese culture. The second image demonstrates \textit{contextual association} where the concept of \textit{artist} is evoked using objects in work-related contexts. The third image shows \textit{symbolic association} where the concept of \textit{Acting} is evoked using known symbols.

By leveraging such associative reasoning, visual metonymy enables artists or advertisers to convey layered meanings that engage the viewer's prior knowledge and interpretation, making it a powerful tool~\citep{forceville2009nonverbal}.  It is a preferred strategy in advertising to point to product or brand identity using visual metonymy, as it is more persuasive and functional than visual metaphors~\citep{forceville-2020}. In spite of visual metonymy largely dominating everyday multimodal communication, it remains a largely unexplored area within the current landscape of multimodal NLP.

Several research gaps persist: i) Studies in the field of NLP have not explored metonymy in a multimodal setting. ii) There are no existing datasets specifically designed around metonymic imagery, nor benchmarks that evaluate a model’s ability to interpret indirect, conceptually grounded metonymic cues. iii) Generating visually metonymic images presents a significant challenge for current language models. Recent studies have shown that while text-to-image models are effective at generating realistic imagery for concrete words with clear visual referents, they frequently fail when prompted with abstract or inferential terms such as \textit{brave} or \textit{consciousness}~\citep{johny_cant_prompt, i_spy_a_metaphor}---terms which are commonly associated with visual metonymy. 

Addressing these gaps, our work is one of the first to explore visual metonymy in the NLP community. We propose a novel image generation framework that leverages large language models (LLMs) to evoke concepts through visual metonymy. 84.3\% of the images generated using our framework were judged as metonymic by humans, compared to only 41.2\% when using a simple prompt without structured guidance. Using this framework, we introduce ViMET---a benchmark dataset containing 2,000 multiple-choice questions, where each image is paired with a multiple-choice question about the concept it is intended to evoke. A majority of existing multimodal benchmarks assess \textit{what} vision-language models (VLMs) see; focusing on object recognition or caption alignment, rather than \textit{how} they interpret visual information~\cite{li2024surveybenchmarksmultimodallarge}. Our benchmark is designed to evaluate the interpretability and depth of an LLM’s visual understanding. Our experiments show that VLMs can reach up to 65.9\% accuracy on our ViMET dataset, while human scores are at 86.9\%. In summary, our contributions are:

\begin{itemize}[leftmargin=*, topsep=2pt, itemsep=2pt, parsep=0pt]
    \item We present the first computational investigation of visual metonymy, laying a foundation for future work.
    \item We propose a cognitively grounded pipeline for generating visual metonymy images, and release a corpus of 2,000 visual metonymy images with rich metadata to support downstream research.
    \item We introduce \textbf{ViMET}, the \textbf{Vi}sual \textbf{Met}onymy benchmark, consisting of 2,000 multiple-choice questions.
    \item Experiments on our dataset show that VLMs' capability of understanding metonymic imagery is significantly (21\%) lower than humans.
\end{itemize}

\section{Modeling Visual Metonymy: A Semiotic and Cognitive Perspective}

\subsection{Semiotic Theory}

Previous linguistic studies have proposed various theoretical perspectives on visual metonymy~\citep{visual_metonymy, visual_design_grammer}. However, no existing study has captured visual metonymy in a computational model, or defined it in a quantitative sense. In this work, we aim to model visual metonymy using  Charles Sanders Peirce’s theory of signs~\citep{peirce1931collected}, particularly the \textbf{semiotic triad} comprising the \textit{object}, \textit{representamen}, and \textit{interpretant}. 

\begin{figure}[t]
    \centering
    \includegraphics[width=0.98\linewidth]{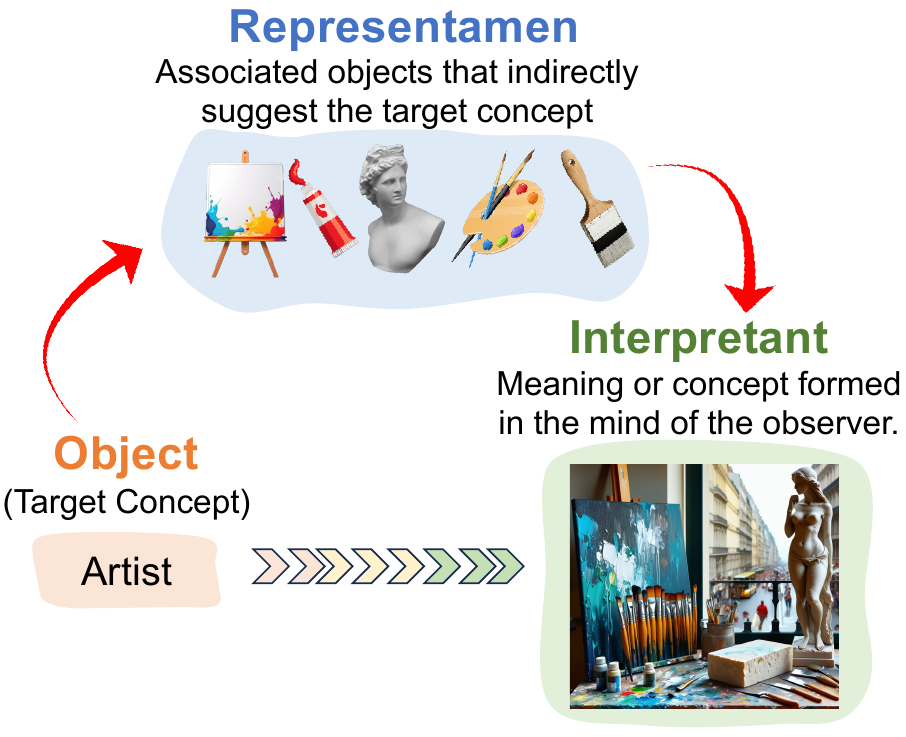}
    \caption{An illustration of the semiotic triad in the context of visual metonymy. The \textit{object} (artist) is indirectly evoked through \textit{representamens} (canvas, color palette, sculpture, brush). The \textit{interpretant} emerges as the viewer infers the concept of an artist from the association of these visual elements without explicit depiction.}
    \label{fig:sem_triad}
\end{figure}

In the semiotic triad, the \textit{object} is the real-world concept, entity, or idea that is being represented. The \textit{representamen} is the perceptible form or sign that stands for the \textit{object}. It is the visible or tangible element that triggers recognition. The \textit{interpretant} is the mental concept or meaning formed by the observer when encountering the \textit{representamen}. This relationship forms a triangular structure---also referred to as the \textit{sign triangle}. The \textit{object} leads to the \textit{representamen}, which in turn leads to the \textit{interpretant}, with the \textit{representamen} acting as a connector between the \textit{object} and the meaning inferred by the observer. This triadic relationship bears a structural resemblance to the Vauquois triangle~\citep{vauquois1968triangle} in machine translation, which maps the translation process from source language to target language via an interlingua or deep structure. Like the semiotic triad, the meaning is mediated through an intermediate representation. 

Figure \ref{fig:sem_triad} depicts the semiotic triad in the context of visual metonymy. The \textit{object} corresponds to the target noun or concept that the image is intended to evoke (e.g., artist). The \textit{representamen} consists of associated objects—visual cues that are not the \textit{object} itself but are closely linked to it through cultural, contextual, or symbolic associations (e.g., canvas, paintings, sculpture, paint brush, and color palette). The \textit{interpretant} occurs when a viewer sees the associated elements (i.e., the \textit{representamens}) and infers the broader concept (artist). Therefore, using the semiotic triad, visual metonymy can be described as a process in which a concept (\textit{object}) is evoked through a related visual cue (\textit{representamen}), prompting the viewer to mentally infer the intended meaning (\textit{interpretant}) without the concept being explicitly depicted. 

The semiotic triad offers a useful cognitive lens for modeling visual metonymy. Prior linguistic and cognitive work has similarly relied on indirect mappings and associative cues~\citep{forceville2009nonverbal, visual_metonymy}, although to our knowledge, this triadic model has not been applied to visual metonymy in computational contexts. The novelty of our work lies here. Our findings suggest that the semiotic triad not only provides a strong theoretical foundation for defining visual metonymy, but also greatly improves the quality of metonymic image generation, discussed in the next section.

\begin{figure}[t]
    \centering
    \includegraphics[width=0.95\linewidth]{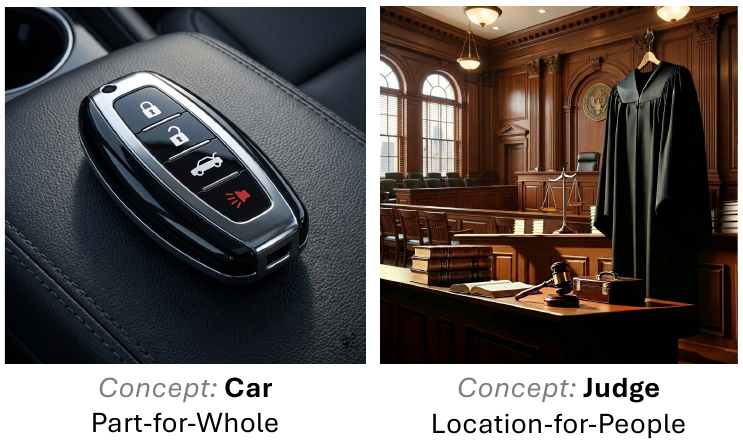}
    \caption{Examples of different visual metonymies.}
    \label{fig:taxonomy}
\end{figure}

\begin{figure*}[t]
    \centering
    \includegraphics[width=0.98\linewidth]{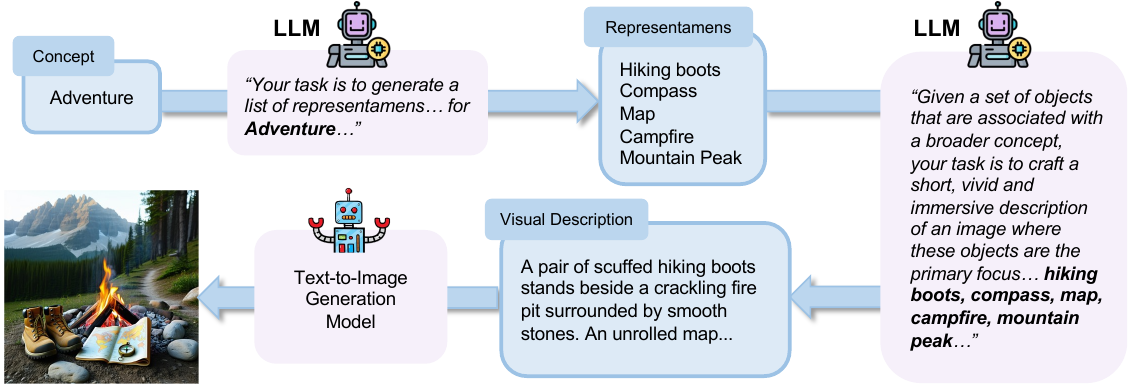}
    \caption{Image generation pipeline. Full prompts are provided in Appendix~\ref{app_prompts_used}.}
    \label{fig:pipeline}
\end{figure*}

\subsection{Association With Textual Metonymy}

Prior linguistic work has explored categories of metonymy in text~\citep{radden}. Recent work by \citet{ghosh2025conmecdatasetmetonymyresolution} adopts six metonymy types: container-for-content, producer-for-product, product-for-producer, location-for-people, possessed-for-possessor, and causer-for-result. Other studies have also identified additional common associations, such as part-for-whole, object-for-action~\citep{kovecses_2010_metaphor}. We observe that visual metonymy can exhibit similar types of associative mappings. Figure~\ref{fig:taxonomy} presents two illustrative examples. The concept of \textit{car} is evoked by \textit{car keys} through part-for-whole metonymy. Similarly, \textit{judge} is evoked by showing the location of a \textit{courtroom}. However, we noticed that in text, a sentence typically instantiates a single metonymic relation anchored by a single cue---often a noun that triggers the intended substitution. For example, in ``\textit{\textbf{Hollywood} released three blockbusters this year},'' the metonymic cue is the noun \textit{Hollywood}, prompting the reader to infer a replacement (e.g., the film industry or its people). 

In contrast, we observed that visual metonymy often simultaneously supports multiple metonymic relations within the same image, drawing on multiple cues distributed across the scene. Since images often contain several objects, tools, and contextual settings at once, each element can independently function as an associative cue pointing to the same target concept via different metonymic mappings. For instance, the concept \textit{judge} in Figure~\ref{fig:taxonomy} can be evoked via location-for-people metonymy, while objects such as a \textit{judge's gown, gavel} and \textit{law books} may additionally cue the concept through object-for-action metonymy. 
This multiplicity makes visual metonymy richer and more compositional than its textual counterpart.

\section{Visual Metonymy Image Generation}

In this section, we propose a novel image generation pipeline with human-in-the-loop evaluation to ensure high-quality visual metonymic images.

\subsection{Image Generation Pipeline}

We propose a three-step approach for generating visually metonymic images from objects. The process involves 1) using an LLM to generate the representamens of the object; 2) using these representamens to generate a visual description of an image with an LLM where the representamens capture the implicit meaning of the broader concept; 3) using a text-to-image model to generate visual metonymy from the LLM-generated visual elaboration. The framework is shown in Figure \ref{fig:pipeline}.

\vspace{3pt}
\noindent \textbf{Generating Representamens.} To generate candidate representamens, we provide Llama 3.1-70B-Instruct~\citep{llama3} with a few illustrative examples~\citep{gpt3_fewshot} and prompt it to produce a set of objects that share a strong cognitive connection to the given noun, ensuring the objects are concrete terms that can be depicted visually. 
We set a temperature and top-p of 0.9, and repetition penalty of 1.1.

\vspace{3pt}
\noindent \textbf{Visual Description using Chain-of-Thought.} We provide the generated representamens from the previous step to Llama 3.1-70B-Instruct and ask the model to produce a visual description of an image. However, a direct prompting strategy often led the model to explicitly mention the concept word, resulting in literal depictions.

To address this, we employed chain-of-thought (CoT) prompting~{\citep{chain_of_thought}}, guiding the model through a reasoning process inspired by the semiotic triad. The CoT prompt includes both the concept word and its associated representamens, and instructs the model to: a) construct a coherent and natural visual description that integrates only those representamens that are sufficient to evoke the concept, rather than forcing all of them into the scene; b) avoid explicitly naming the concept word in the description; and c) incorporate visual compositional elements such as tone, lighting, and color palette to reinforce the intended concept.

\begin{figure*}[t]
    \centering
    \includegraphics[width=0.98\linewidth]{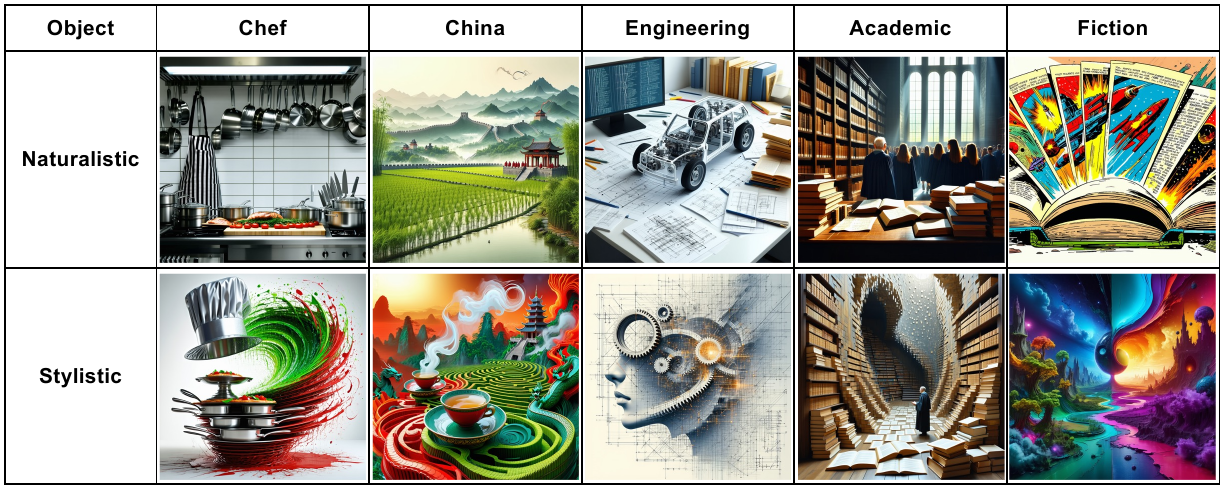}
    \caption{Examples of visual metonymy images generated by our pipeline.}
    \label{fig:dataset_examples}
\end{figure*}

\begin{figure}[t]
    \centering
    \includegraphics[width=0.99\linewidth]{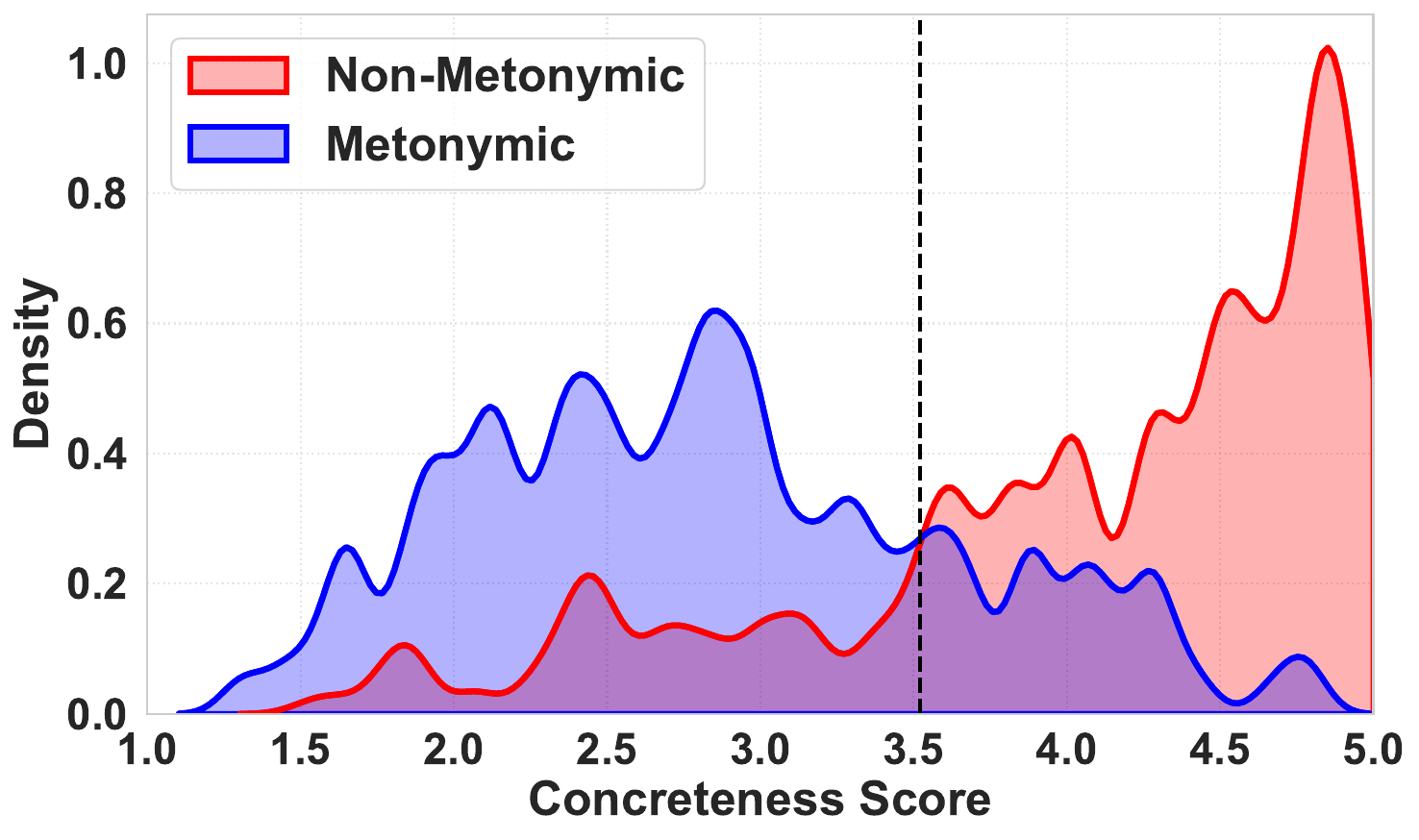}
    \caption{Density vs concreteness rating for the 1,000 annotated images.}
    \label{fig:conc_density}
\end{figure}

We included a fail-safe mechanism here that checks for the presence of the concept word in the output and re-generates the description if detected. Furthermore, CoT prompting allowed us to control the stylistic tone of the descriptions, ensuring that multiple stylistic interpretations of the same concept are captured, providing a richer and more varied dataset. For each concept word, we generated two distinct descriptions corresponding to different styles: \textbf{Naturalistic} (compositional and grounded in reality) and \textbf{Stylistic} (non-compositional and visually creative). We do this to evaluate how visual style impacts image understanding~\citep{peng2017visdavisualdomainadaptation, geirhos2018imagenettrained}. We set a temperature and top-p of 0.9, and repetition penalty of 1.1. 

\vspace{3pt}
\noindent \textbf{Generating Visual Metonymic Images.} Finally, we feed the visual descriptions generated by Llama into Stable Diffusion-3.5-Large~\citep{stable_diffusion}. We use 35 inference steps, with a guidance scale of 7.5, which strikes a balance between adherence to the prompt and creative variation.

\subsection{Human Feedback Based Concept Filtering}

To generate metonymic images, we start with collecting concepts from WordNet~\citep{wordnet}. Since textual metonymy typically operates through nouns, we restrict our focus to nouns as the concepts. Specifically, we use the 26 semantic common noun classes called \textit{supersense}, proposed by \citet{wordnet_supersense}. We randomly select nouns from each of these supersense classes to generate metonymic images using our pipeline. We then sampled 1,000 images and recruited human annotators to identify if the image evokes the idea of the concept without explicitly showing it. Full annotation details are provided in Appendix~\ref{annotation_guidelines}. The raw agreement between the annotators was measured to be 89.2\%, calculated as the proportion of identical labels out of the total instances.

Based on the human feedback, we identified that it is not easy to represent certain concepts metonymically through visual cues. For example, concrete concepts such as \textit{table} or \textit{chair} are difficult to depict metonymically without directly illustrating the concept itself, limiting the potential for associative interpretation. Drawing from the annotation results, we introduce two criteria to filter out concepts that have a lower likelihood of producing successful metonymic images. These criteria are as follows:

\vspace{3pt}
\noindent\textbf{Concreteness Score Based Filtering.} To systematically identify concepts that are more likely to support visual metonymy, we leverage concreteness scores from \citet{concreteness}, which provide concreteness ratings for approximately 19,056 nouns ranging from 1 (highly abstract) to 5 (highly concrete).  Figure \ref{fig:conc_density} illustrates the distribution of concreteness scores for metonymic and non-metonymic instances in the annotated samples. The x-axis denotes the concreteness score, while the y-axis reflects the estimated density, representing a smoothed histogram. At a concreteness value of approximately 3.5, the red curve (non-metonymic) overtakes the blue curve (metonymic), suggesting that concepts with higher concreteness are more likely to be interpreted as non-metonymic. Using this observation, we only use concepts with a concreteness score below 3.5.

\vspace{3pt}
\noindent \textbf{WordNet Category Based Filtering.} Based on the annotated samples, we retain only those WordNet supersense labels in which more than 60\% of the images were annotated as metonymic. The categories that met this criterion include: \textit{act, attribute, cognition, communication, event, feeling, group, location, motive, person, possession, process, state}, and \textit{time}. More details are provided in Appendix~\ref{appendix_wordnet_categories}.

In summary, we only select the concepts that belong to these categories and have a concreteness rating less than 3.5.

\begin{table}
\centering
\resizebox{0.95\linewidth}{!}{
\begin{tabular}{lcc}
\toprule
 & \textbf{Our Pipeline} & \textbf{General} \\ 
\midrule
Human Evaluation & \textbf{84.3\%} & 41.2\%  \\
\bottomrule
\end{tabular}}
\caption{Percentage of metonymic images in a sample of 1,000 according to human evaluation, comparing images generated by our pipeline against a general prompting using only the concept word.}
\label{tab:vimet_dataset_eval}
\end{table}

\subsection{Pipeline Evaluation}
\label{pipeline_evaluation}
We apply two filtering criteria to the full set of 19,056 nouns from WordNet, resulting in a subset of 2,077 concept words. Our image generation pipeline is then applied to each of these concepts, producing two images per concept and generating a total of 4,154 images. Figure~\ref{fig:dataset_examples} shows some visual metonymic images created by our pipeline in naturalistic and stylistic tone for the same concept. To evaluate the quality of the images generated by our pipeline, we conduct human annotation. We sample 1,000 images and ask the annotators to independently label each image as metonymic or non-metonymic based on previously established guidelines. We compare images generated by our pipeline with the images generated by only providing the concept word using general prompting to Stable Diffusion. Table~\ref{tab:vimet_dataset_eval} shows the evaluation results. The agreement between annotators was 92.4\%. \textbf{84.3\%} of the images generated by our pipeline were labeled as metonymic, compared to only \textbf{41.2\%} of those produced with a simple prompt. This highlights the effectiveness of our pipeline in producing non-literal, conceptually grounded visual representations, while demonstrating the struggle of image generation models evoking a concept without a structured approach. More examples are provided in Appendix~\ref{direct_comparison_appendix}. 

\section{ViMET: Our New Dataset}

Visual metonymy understanding requires models to go beyond surface-level perception and engage in conceptual association through non-literal interpretation. To this end, we leverage our framework to evaluate cognitive reasoning capabilities of VLMs by annotating more images and selecting 1,000 concept words labeled by humans to produce metonymic images across both visual styles (naturalistic and stylistic), creating the \textbf{ViMET} dataset. The dataset is composed of 2,000 images across 1,000 concept words as gold labels, each image accompanied by a multiple-choice question. In each instance, the model is presented with an image and four candidate concepts, and must select the option that best aligns with the concept the image implicitly conveys. 

\vspace{3pt}
\noindent \textbf{Creating the Distractors.} To construct plausible false options for the multiple-choice questions (MCQ), we combine both visual and semantic similarity. We first compute the CLIP embedding~\citep{clip} of the target concept's image and retrieve the concept words whose corresponding images are most visually similar from the full set of generated images. This ensures the distractors pose a visual challenge. We then use the `\texttt{/r/RelatedTo}' relation from ConceptNet~\citep{conceptnet} to get semantic distractors that are related to the concept. 

\begin{figure}[t]
    \centering
    \includegraphics[width=0.99\linewidth]{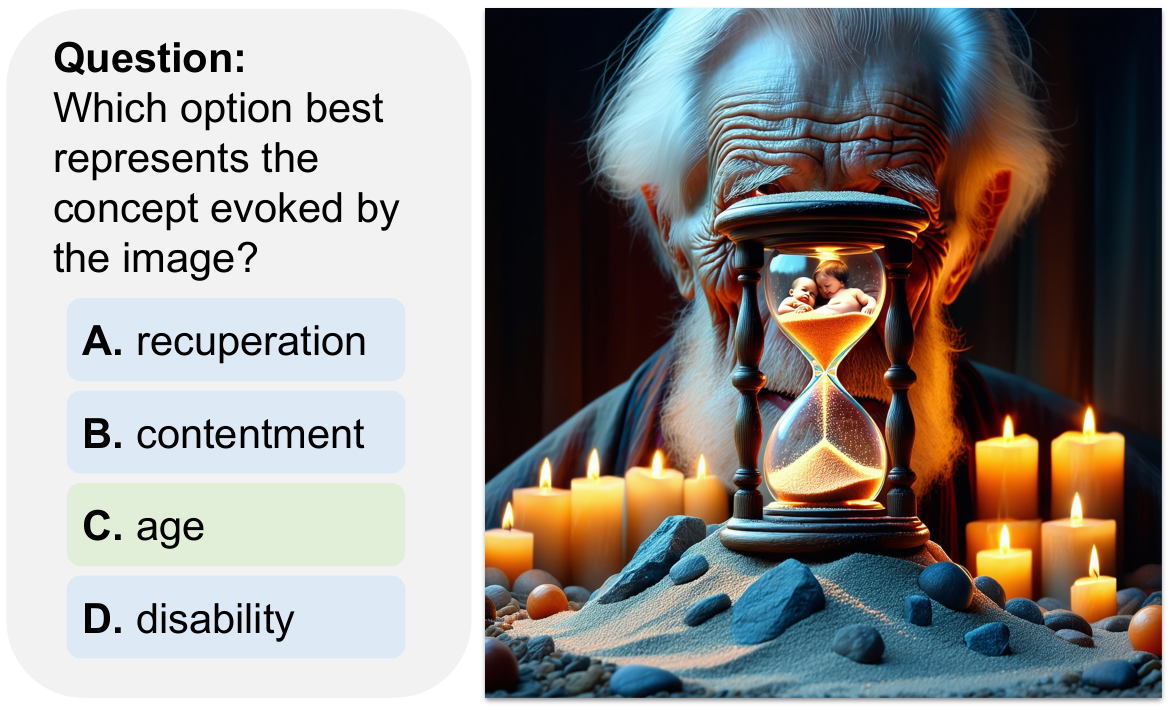}
    \caption{Example of a question from ViMET dataset.}
    \label{fig:qa_example}
\end{figure}

However, this introduces potential ambiguity, as some candidate distractors may be synonyms or semantically too similar to the target concept. To effectively manage and precisely control this ambiguity, we implement three complementary filtering strategies: 1) removing synonyms using ConceptNet via the \texttt{/r/Synonym} relation; 2) semantic similarity filtering using BERT embeddings~\citep{devlin-etal-2019-bert} to exclude excessively similar terms based on cosine similarity; 3) 2-step relational distance filtering in the ConceptNet graph~\citep{semantic_distance}, where distractors are selected if they are indirectly related to the target concept through an intermediate concept but not directly connected. These filtering mechanisms ensure that the distractors are both semantically and visually proximate to the target concept, increasing the overall difficulty while preserving a controlled level of ambiguity.

Figure \ref{fig:qa_example} shows an example of a question from the ViMET dataset. The correct answer, \textit{age}, is an abstract term and is evoked through symbolic elements such as the elderly man, hourglass, and child within the sand—together suggesting the passage of time. Distractors like \textit{recuperation}, \textit{contentment}, and \textit{disability} may seem related but lack visual or conceptual grounding in the image. This highlights how our dataset challenges models to go beyond surface recognition and engage in cognitive association to identify metonymic meaning. 

\begin{table}
    \centering
    \resizebox{0.9\linewidth}{!}{
    \begin{tabular}{lccc}
    \toprule
    \textbf{Model} & \textbf{Nat.} & \textbf{Styl.} & \makecell{\textbf{Overall}\\ \textbf{Acc.}} \\
    \midrule
    Llama 3.2 11B & 61.5 & 60.8 & 61.2 \\
    Llama 3.2 90B & 61.5 & 63.9 & 62.7 \\
    Llama 4 Scout & 62.8 & 63.5 & 63.2 \\
    \midrule
    InternVL3 8B & 62.2 & 63.6 & 62.9 \\
    InternVL3 78B & 65.4 & 66.4 & 65.9 \\
    \midrule
    Qwen2.5-VL 7B & 64.8 & 62.6 & 63.7 \\
    Qwen2.5-VL 72B & 66.4 & 64.4 & 65.4 \\
    \midrule
    Gemini 2.5 Flash & 62.3 & 62.9 & 62.6 \\
    Gemini 2.5 Pro & 66.2 & 64.2 & 65.2 \\
    \midrule
    Human Performance & \textbf{85.6} & \textbf{88.1} & \textbf{86.9} \\
    \bottomrule
    \end{tabular}}
    \caption{Accuracy of VLMs and humans on our ViMET dataset. \textbf{Nat.}: 1,000 images rendered in Naturalistic style. \textbf{Styl.}: 1,000 images rendered in Stylistic style. \textbf{Overall Acc.}: Accuracy on all 2,000 samples.}
    \label{tab:qa_task}
\end{table}

\vspace{3pt}
\noindent \textbf{ViMET Dataset Results.} Table \ref{tab:qa_task} shows the performance of VLMs and humans on the ViMET dataset. We evaluate four VLM families: Llama~\citep{llama3,llama4meta}, InternVL3~\citep{zhu2025internvl3exploringadvancedtraining}, Qwen2.5-VL ~\citep{qwen2025qwen25technicalreport}, and Gemini 2.5~\citep{gemini25}. Naturalistic images are compositional and grounded in reality, whereas Stylistic ones are non-compositional and visually creative. InternVL3 78B achieves the highest accuracy on stylistic images, while Qwen2.5-VL 72B performs best on naturalistic images. The overall best model (InternVL3 78B) falls notably short of human performance by 21.0\%. This disparity highlights not only the limitations of current VLMs in handling cognitive associations, but also underscores the inherent difficulty of our benchmark, which requires models to interpret subtle, non-literal visual cues to infer conceptual meaning.

\section{Analysis \& Discussion}

\noindent \textbf{Generalization of our Framework.} \label{generalizability}  To assess the generalizability of our visual metonymy image generation framework across different LLMs: Llama3, Qwen3~\citep{yang2025qwen3technicalreport}, GPT-OSS~\citep{openai2025gptoss120bgptoss20bmodel}, Gemini, and GPT-4o~\citep{openai2024gpt4technicalreport}; and image generation models: Stable Diffusion, GPT-4o, Janus-7B-Pro~\citep{janus_7b_pro}, and Qwen-Image~\citep{qwen2025qwen25technicalreport}, we sampled 100 concepts and generated images using various model combinations. Human annotators then judged whether the outputs were metonymic, following the same annotation guidelines described earlier. Table~\ref{table:generalizability} shows the results. Stable Diffusion (SD), GPT-4o, and Qwen-Image achieve comparable performance. Janus-7B-Pro performs slightly worse, which we attribute to its limited 384x384 output resolution~\citep{janus_7b_pro}, making the details of the image less perceptible for annotators. Overall, these findings suggest a key takeaway---our framework generalizes across a range of text and image generation models, demonstrating its flexibility for future implementations. 

\begingroup
\begin{table}[t]
   \centering
   \resizebox{0.98\linewidth}{!}{
   \begin{tabular}{lcccc}
   \toprule
   LLM & \makecell{SD}& \makecell{GPT\\4o} & \makecell{Janus\\Pro 7B} & \makecell{Qwen\\Image} \\
   \midrule
   *Llama-3.1-70B & 85 & 85 & 77 & 83\\
   *Qwen3-30B & 84 & 86 & 76 & 84\\
   GPT-OSS-20B & 81 & 81 & 75 & 80\\
   Gemini 2.5 Flash & 84 & 84 & 75 & 81\\
   GPT-4o & 87 & 82 & 77 & 87\\
   \bottomrule
   \end{tabular}
   }
\caption{Number of images annotated as metonymic by human judges in a sample of 100 concepts. The representamens and visual descriptions are generated by the LLMs (* denotes Instruct version of the model). The image is generated by the image generation models. For ViMET, we selected Llama-3.1-70B-Instruct and Stable Diffusion-3.5-large as the primary backbone model.}
\label{table:generalizability}
\end{table}
\endgroup

\vspace{3pt}
\noindent \textbf{Analyzing Conceptual Association.} \label{conceptual_association_analysis} To provide further insight into the conceptual association capabilities of VLMs, we sampled 250 responses from the ViMET dataset and categorized them based on the type of metonymic association---cultural, contextual, and symbolic. Table~\ref{tab:association} presents the percentage of correct responses for each category. A consistent trend emerges: symbolic associations yield the highest accuracy suggesting that they are often more visually distinct, followed by cultural, while contextual associations result in the lowest performance, indicating they are harder to infer. This pattern can be attributed to the nature of the associations. Symbolic associations typically involve a well-defined and culturally reinforced symbol that directly maps to a concept (e.g., \textit{dove} representing \textit{peace}, or \textit{snake} symbolizing \textit{betrayal}). Once the symbol is recognized, the meaning of the image becomes relatively easy to interpret. Cultural associations function similarly, relying on shared cultural knowledge to bridge the gap between the image and the concept. In contrast, contextual associations tend to be more ambiguous. The representamens in these cases co-occur with the concept in real-world scenarios, but do not point to it uniquely. For instance, objects like a \textit{pointer, projector}, and \textit{whiteboard} might suggest \textit{teaching}, but they could just as easily relate to \textit{presentations} or \textit{seminars}, making the intended concept more difficult to infer with certainty. Examples of each type of association are given in Appendix~\ref{appendix_cultural_association}-\ref{appendix_symbolic_association}.

\begin{table}
    \centering
    \resizebox{0.95\linewidth}{!}{
    \begin{tabular}{lccc}
    \toprule
     & Cultural & Contextual & Symbolic \\ 
    \midrule
    VLMs & 66.6 & 54.5 & 76.3  \\
    Humans & 88.3 & 75.2 & 92.1  \\
    \bottomrule
    \end{tabular}}
    \caption{Percentage of correct responses of VLMs in the ViMET for different modes of concept association. }
    \label{tab:association}
\end{table}

\vspace{3pt}
\noindent \textbf{Visual Metonymy vs. Metaphor.} \label{discussion} Prior linguistic studies~\citep{lakoff1980, dirven2002metonymy} argued that unlike metaphor, which maps meaning across different conceptual domains,  metonymy operates within a single domain by invoking a concept through its association with another. This observation is also reflected in the visual modality, where the work of \citet{gutierrez-etal-2016-literal} and \citet{i_spy_a_metaphor} grounds metaphor in cross-domain compositional structure. Corroborating observations made by previous linguistic studies, our work shows that visual metonymy relies on cognitive association within a single domain, making it less dependent on visual form and more on the viewer’s ability to connect what is seen to what is implied. 

Prior studies~\citep{gal_2019, i_spy_a_metaphor} revealed that metaphor is best understood as visual material rather than a purely conceptual phenomenon. It emerges primarily through the compositionality of the image, e.g., visual form, or material arrangement that shape how meaning is perceived. Interestingly, our findings suggest that visual metonymy operates more as a conceptual phenomenon, where meaning arises through the viewer’s interpretation of the image. As illustrated in Figure~\ref{fig:dataset_examples}, the naturalistic and stylistic images in our ViMET dataset often exhibit significantly different visual compositions for the same concept word, yet are united through their conceptual linkage in the viewer’s mind. 

\begin{figure}
    \centering
    \includegraphics[width=0.98\linewidth]{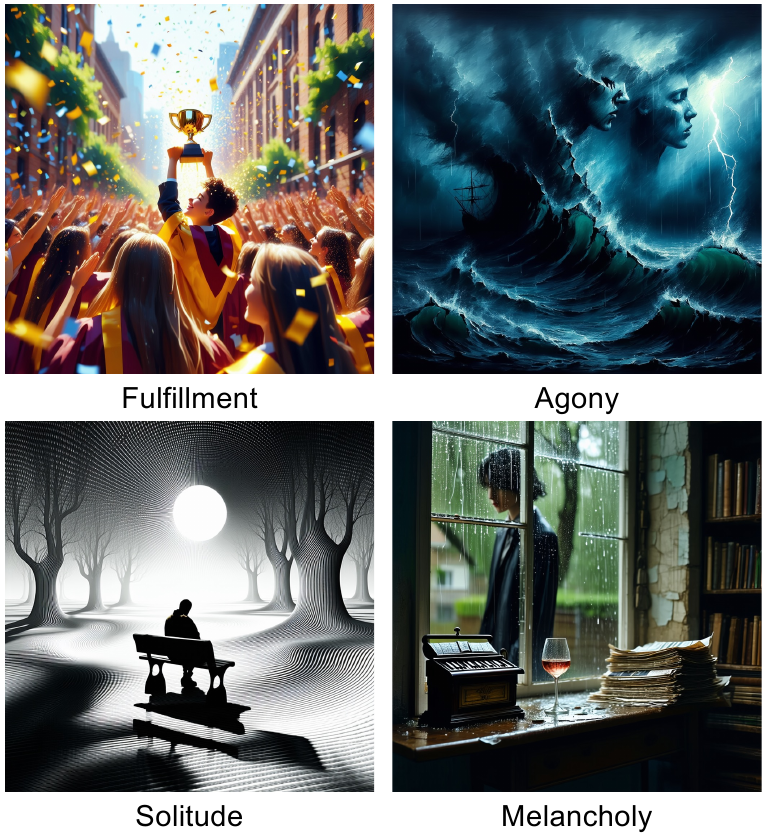}
    \caption{Examples of metonymic images with different visual tones to evoke the concept.}
    \label{fig:tone}
\end{figure}

\vspace{3pt}
\noindent \textbf{Evoking Concept Through Visual Tone.} While the presence of appropriate representamens is central to evoking a concept metonymically, we find that for certain abstract concepts, tone and mood play a critical role. In our dataset, we observe that many images successfully evoke a concept not just through objects, but through carefully crafted atmospheres---such as color palette, lighting, and mood---that resonate with the intended meaning. This is particularly important for abstract or emotional concepts, where literal objects alone may be insufficient. The strength of our image generation pipeline lies in part in the use of chain-of-thought prompting, which encourages the model to produce descriptions that naturally incorporate the appropriate emotional and conceptual tone, resulting in more nuanced and expressive visual outputs. Figure~\ref{fig:tone} shows some examples of such cases.

\section{Related Work}

\textbf{Metonymy in NLP.} Metonymy has long been recognized as a significant concept across multiple disciplines, including philosophy, linguistics, and psychology~\citep{jakobson1956, dirven2002metonymy}. Building on cognitive theories by \citet{lakoff1980}, metonymy is understood as a conceptual mapping where one entity (e.g., \textit{dish}) stands for another related entity (e.g., \textit{food}). Unlike metaphor, metonymy is often approached as a cognitive and pragmatic phenomenon, rather than purely linguistic terms~\citep{radden, maria2015metonymy}. \citet{papafragou1996metonymy} argues that metonymy is a variety of interpretive use, specifically an instance of name introduction. In NLP, metonymy is often studied for named entities, and has been framed as a classification problem known as metonymy resolution~\citep{markert-nissim-2002-metonymy}. Datasets across locations and organizations such as the SemEval 2007 Shared Task 8~\citep{markert-nissim-2007-semeval}, RelocaR~\citep{gritta-etal-2017-vancouver}, WIMCOR~\citep{alex-mathews-strube-2020-large}, and ConMeC~\citep{ghosh2025conmecdatasetmetonymyresolution} have provided valuable resources for developing NLP systems to identify textual metonymy. More recently, \citet{ghosh2025conmecdatasetmetonymyresolution} introduced a metonymy dataset that focuses on common nouns. Despite this progress, existing computational work largely focuses on language, with visual metonymy unexplored.

\vspace{3pt}
\noindent \textbf{Visual Reasoning.} Our work connects to the broader area of visual reasoning, where the field has developed numerous benchmarks to evaluate multimodal understanding~\citep{VQA, zellers2019recognition, hudson2019gqa, marino2019ok}. More recently, benchmarks such as MVP-Bench~\citep{li2024mvpbenchlargevisionlanguagemodels}, MMDocBench~\citep{zhu2024mmdocbenchbenchmarkinglargevisionlanguage} and CulturalVQA~\citep{nayak2024benchmarkingvisionlanguagemodels} aim to evaluate the deeper understanding capabilities of VLMs across diverse domains and settings. A consistent finding is that VLMs struggle more with abstract concepts than with concrete ones in visual tasks~\citep{tater-etal-2025-absvis, talon2025seeingabstracttranslatingabstract}. Several works have explored how multimodal models see abstract concepts in images~\citep{gunther2022images, pezzelle-etal-2021-word}, as well as how abstract and concrete concepts differ in their visual realizations~\citep{khaliq-etal-2024-comparison, tater-etal-2024-unveiling}. This difficulty with abstract concepts extends to image generation as well, with popular text-to-image generators struggling to visualize abstract terms, motivating studies to develop frameworks to improve abstract concept visualization~\citep{Fan_2024, liao2024}.

\vspace{3pt}
\noindent \textbf{Visual Representation of Figurative Expressions.} Figurative language such as metaphors, metonymy, and idioms poses a special challenge for multimodal NLP, since understanding non-literal expressions alongside images demands commonsense and cultural knowledge~\citep{yosef-etal-2023-irfl}, as well as the ability to map between abstract and concrete domains~\citep{hill-korhonen-2014-learning}. Metaphors or idioms in captions often rely on unconventional, implied meanings that go beyond the literal visual cues, making them harder for models to grasp~\citep{akula}. To address this, researchers have established dedicated benchmarks, such as the IRFL dataset that pairs idioms, metaphors and similes~\citep{yosef-etal-2023-irfl}, the SemEval-2025 AdMIRe shared task~\citep{pickard-etal-2025-semeval} that introduced multilingual multimodal idiomaticity benchmarks. MultiMET~\citep{zhang-etal-2021-multimet} curates text-image pairs with annotations for metaphor presence and ACD~\citep{rajakumar-kalarani-etal-2024-unveiling} introduces a metaphor captioning dataset. Other metaphor datasets include V-Flute~\citep{saakyan2025understandingfigurativemeaningexplainable}, MetaCLUE~\citep{akula}, MET-Meme~\citep{met-meme}, and Hummus~\citep{tong2025hummusdatasethumorousmultimodal} for humorous metaphor understanding in advertising images, internet memes and humor. \citet{kundu-etal-2025-looking} showed that VLMs significantly lag behind humans in understanding visual metaphors. Studies have also explored generating visual metaphors, using LLM prompting with human feedback to produce highly-compositional outputs~\citep{i_spy_a_metaphor} and multi-faceted reward signals for iterative or RL-based refinement~\citep{koushik2025mindseyemultifacetedreward}. A closely aligned semiotics-based approach to ours is presented by \citet{kruk-etal-2023-impressions}, who propose a dataset for studying how visual elements and design choices systematically evoke particular emotions, thoughts, and beliefs. While prior studies have explored visual metonymy through a theoretical lens~\citep{forceville2009nonverbal, visual_metonymy}, these works remain limited to qualitative analysis without benchmarks or computational application. To our knowledge, we are the first to introduce a visual metonymy dataset, and explore this phenomenon in the field of NLP.

\section{Conclusion}

In this work, we explored the concept of visual metonymy in the computational domain. We presented a novel image generation pipeline grounded in semiotic theory that enhances visual metonymic imagery---addressing the limitations of existing vision systems in representing abstract concepts. We introduced ViMET, a benchmark dataset to evaluate models’ ability to perform cognitive association and non-literal interpretation. Our results revealed that current vision-language models struggle to identify the underlying concept of an image, with performance approximately 20\% below human capability. We hope this work sheds light on the importance of visual metonymy, and our ViMET dataset will contribute to the cognitive visual reasoning in VLMs.

\section*{Limitations}

While our image generation pipeline---grounded in the semiotic triad---substantially improves the quality of metonymic image representation, we do not independently evaluate the effectiveness of each intermediate step in the pipeline. In particular, assessing the quality of the generated representamens and the fidelity of the visual descriptions constructed by the LLM based on those representamens is non-trivial and would require targeted human evaluation. Although informal inspection suggests these intermediate outputs are meaningful and coherent, we lack a systematic, scientific assessment of their individual contributions to the overall metonymic quality. Future work could introduce step-wise human evaluation of each intermediate step.

In our framework, we use LLMs to generate representamens that enable the viewer to infer the concept through association. Since these associations are derived from the LLM’s learned knowledge, they are inherently constrained by the model’s training data. As a result, if a culturally or contextually significant association is not captured within the LLM’s knowledge base, it may be omitted—leading to gaps in the representational coverage for certain concepts.

Visual metonymy is inherently open to interpretation. An image can evoke multiple plausible concepts depending on a viewer’s knowledge, cultural context, and attention to different visual cues. Unlike many object-centric vision tasks with a single unambiguous label, metonymic meaning is often conveyed indirectly through associations, making strict ``one correct answer'' evaluation challenging. Although we mitigate this issue through careful prompt design, metadata, and human validation, some degree of ambiguity remains unavoidable.

\section*{Ethical Considerations}

Our work involves the use of large-scale text-to-image generation models, which may raise concerns regarding intellectual property and copyright. We clarify that all outputs were used strictly in the context of academic research and are not intended for commercial use. We do not claim authorship or originality of individual image outputs. 

A key ethical concern in generating metonymic imagery is the risk of reinforcing or amplifying stereotypical associations. For example, generative models often depict cultural or national concepts (e.g., Japan, Brazil) through reductive or stereotypical imagery. We explicitly acknowledge these limitations and caution that metonymic image generation is inherently prone to such representational tendency. We also emphasize that the task of visual metonymy generation or identification itself highlights an area of ethical concern: the tendency of humans or models to reproduce cultural stereotypes and biased representations. We therefore include this discussion not only as a limitation of our study but also as a call for future research to develop frameworks that systematically detect, quantify, and mitigate representational harms in multimodal generation systems. 

Lastly, visual metonymy generation can also result in violent, graphic, or otherwise uncomfortable imagery depending on the concept provided to the model. We carefully curated our dataset by manually removing such outputs, ensuring that both the ViMET dataset and the publicly available image repository are clean and safe for use by the NLP community. Further details of this filtering process are provided in Appendix~\ref{toxicity}.

\section*{Acknowledgments}

We thank the CincyNLP group for their helpful comments and the anonymous EACL reviewers for their valuable feedback and suggestions.

\bibliography{custom}

\clearpage
\newpage

\appendix


\section{Analyzing Representamens}
\label{analyizing_rep}

\begin{figure}[h]
    \centering
    \includegraphics[width=0.98\linewidth]{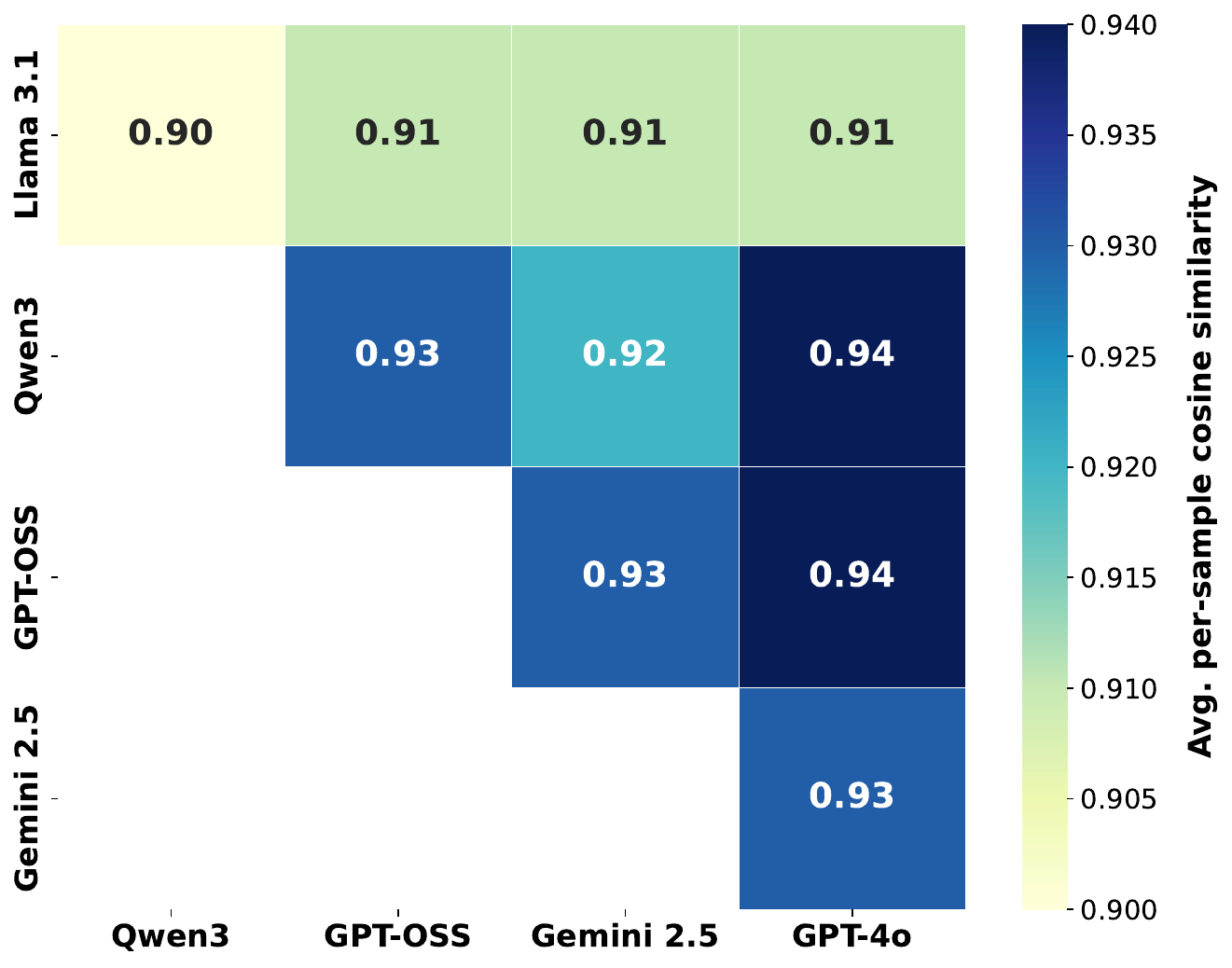}
    \caption{Average cosine similarities of representamens generated by all models for one sample.}
    \label{fig:rep_heatmap}
\end{figure}

We use 4 different LLMs to generate representamens and visual descriptions in Section~\ref{generalizability} to showcase the generalizable nature of our framework. Our goal in this section is to study how similar the generated representamens for a concept are across different LLMs. We encode the representamens using BERT embeddings and apply a greedy matching method to compute the average similarity of all representamens generated for a single concept across models. Figure~\ref{fig:rep_heatmap} shows the results. We observe that the similarity is consistently high across all LLMs, indicating that different models converge on highly overlapping sets of representamens for the same concept. This suggests that our framework is robust to the choice of underlying LLM and that the semantic space of generated representamens remains stable across model variations.

\section{Error Analysis}
\label{error_analysis}

\subsection{Image Generation}

\begin{figure}[t]
    \centering
    \includegraphics[width=0.95\linewidth]{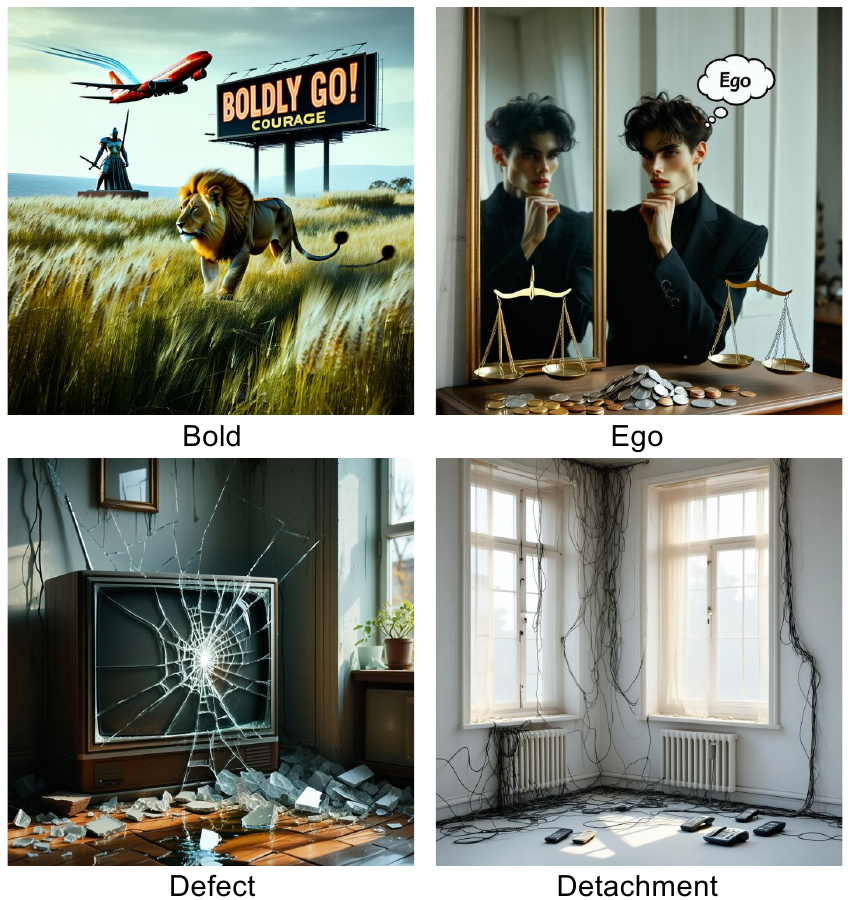}
    \caption{Examples of images generated that were labeled as non-metonymic by human judges.}
    \label{fig:error_gen}
\end{figure}

Figure~\ref{fig:error_gen} shows some examples of images that were annotated as non-metonymic by human judges. Two primary reasons were identified as the cause as to why the generated images were non-metonymic. Firstly, the concept word was explicitly shown in the image either visually or by text, rendering it non-metonymic. The images \textit{Bold} and \textit{Ego} from figure~\ref{fig:error_gen} are two such examples where the concept is shown by text. To identify this was caused by the image generation model or the LLM, we looked into the generated representamens and visual descriptions by the LLMs. According to our observation, these errors were primarily caused by the LLM failing to follow the prompt instructions---generating the concept word in the visual description. For example, for the concept \textit{January}, the LLM produced the phrase ``\textit{...a calendar showing the first month of the year...},''' which resulted in an image that visibly depicted the word itself. Another interesting scenario is when the LLM follows instruction by not explicitly stating the concept, but mentions the concept indirectly. For example, the visual description for the concept \textit{Doctor} stated \textit{``...a man standing with a stethoscope around his neck....''}, and the resulting image from this description explicitly showed a doctor rather than evoking the concept. Such cases show the difficulty to fully suppress the concept word in the visual description. Another common reason why the images were labeled as non-metonymic was because the human judges felt like the image does not evoke the concept of the target word. The images \textit{Defect} and \textit{Detachment} from figure~\ref{fig:error_gen} are two such examples. Such concepts were primarily identified to lack common cultural and contextual representamens.

\subsection{ViMET Dataset Error Cases}

\begin{figure*}[t]
    \centering
    \includegraphics[width=0.98\linewidth]{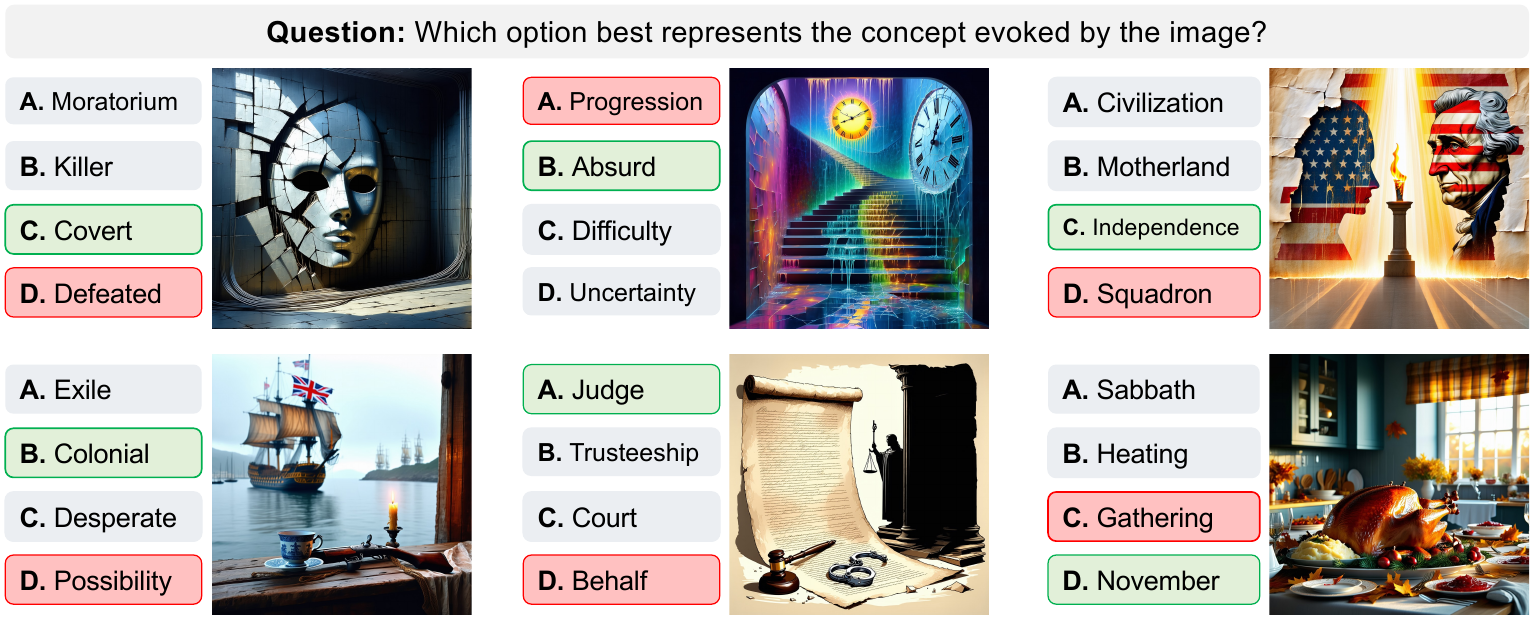}
    \caption{Examples of incorrect responses by VLMs in the ViMET dataset. The green box indicates the correct response, and the red box is the incorrect response given by the VLM.}
    \label{fig:vimet_error}
\end{figure*}

Figure~\ref{fig:vimet_error} shows some of the errors made by the VLMs in the ViMET dataset.

\section{Toxicity, Bias and Violence}
\label{toxicity}

While generating images from the concept words sourced from \citet{concreteness}, we encountered several instances of violent or toxic imagery. This was expected in cases where the concept word itself was inherently graphic, violent or harmful. However, we also observed unexpected occurrences where seemingly neutral words resulted in violent depictions---often due to the representamen objects associated with those concepts having implicit violent associations. We also observed bias in some generated images, particularly with Stable Diffusion, which occasionally associated public figures with negative or stereotypical representations. We observed these biases were not present in the visual prompts generated by Llama, indicating that the issue arose during the image generation stage rather than the textual input. To ensure the integrity and ethical safety of our ViMET benchmark, we manually reviewed and removed all instances containing graphic violence, toxicity, or harmful social biases during final round of annotation. The annotators were instructed to flag potentially graphic or problematic imagery. Out of the 2,077 concept words filtered in Section~\ref{pipeline_evaluation}, 14 concepts were removed due to violent/graphic imagery and 10 were removed due to negative cultural or public figure association.

\section{Dependence on Representamens}

Our core pipeline depends on an object having certain cognitive associations through which metonymy can be represented. Certain concepts naturally lend themselves to metonymic representation through rich cultural or contextual associations, while others do not. For instance, an animal such as a \textit{dog} can be represented metonymically through objects like \textit{whiskers on a sofa, paw marks on a carpet, scratch marks, leashes}, and \textit{pet toys}, all deeply embedded in everyday cultural contexts of domestic life. In contrast, animals like an \textit{ox} or a \textit{donkey} are much harder to evoke through visual representamens, as they hold limited presence in contemporary daily life and thus lack strong, easily recognizable associative imagery. In such cases, the generated representamens are generic and not well connected to the concept, and Llama provides a description of the concept in the visual prompt as it struggles to generate a prompt that conveys the metonymic nature of the image by naturally integrating the representamens. This results in the image having literal depiction of the concept, rendering it non-metonymic. In other cases where the concept is not explicitly depicted, annotators find a hard time connecting the concept to the image, due to the generic nature of the representamens. 

\section{WordNet Categories}
\label{appendix_wordnet_categories}

\begin{table}
  \centering
  \resizebox{0.95\linewidth}{!}{
  \begin{tabular}{lc}
    \toprule
    \textbf{Category} & \textbf{Met. \%} \\
    \midrule
    act     & 79.5           \\
    \tr{animal}     & \tr{13.4}           \\
    \tr{artifact}     & \tr{26.6}           \\
    attribute     & 64.4           \\
    \tr{body}      & \tr{0.0}            \\
    cognition     & 72.3           \\
    communication     & 63.7           \\
    event     & 72.4           \\
    feeling     & 62.8           \\
    \tr{food}     & \tr{9.4}           \\
    group     & 60.2           \\
    location     & 64.5           \\
    motive & 75.0 \\ \bottomrule
  \end{tabular}
  \begin{tabular}{lc}
    \toprule
    \textbf{Category} & \textbf{Met. \%} \\
    \midrule
    \tr{object}     & \tr{30.4}           \\
    person     & 70.6           \\
    \tr{phenomenon}     & \tr{33.3}           \\
    \tr{plant}     & \tr{14.2}           \\
    possession      & 65.6            \\
    process     & 90.9           \\
    \tr{quantity}     & \tr{14.3}           \\
    \tr{relation}     & \tr{34.6}           \\
    \tr{shape}     & \tr{18.4}          \\
    state     & 82.3           \\
    \tr{substance}     & \tr{33.3}           \\
    time     & 75.0           \\
    \- & \- \\
    \bottomrule
  \end{tabular}
  }
  \caption{Table showing percentage of metonymic examples after human annotation from each category.}
  \label{tab:wordnet_score}
\end{table}

Table \ref{tab:wordnet_score} shows the percentage of metonymic images produced by each WordNet supersense category according to human evaluation. The categories marked in red were discarded before sampling concepts for the ViMET dataset. The primary reasons these words were deemed to produce non-metonymic imagery were: a) Highly concrete terms: The words in these categories were primarily concrete words, with most of them having average concreteness scores above 4.0, making it difficult to represent the concept without visually depicting it. b) Lack of representamens: Most of the words in these categories lack recognizable associations due to the rarity of these terms. While well known animals, foods, or plants could be depicted using representamens (e.g., \textit{pizza} was depicted using Italian architecture, oven, seasonings, all without explicitly showing the food item), most words lacked such well-known associations, resulting in  the annotators failing to connect the concept word to the image.

\subsection{Predicting the Concept Word}

\begin{figure}[h]
    \centering
    \includegraphics[width=0.95\linewidth]{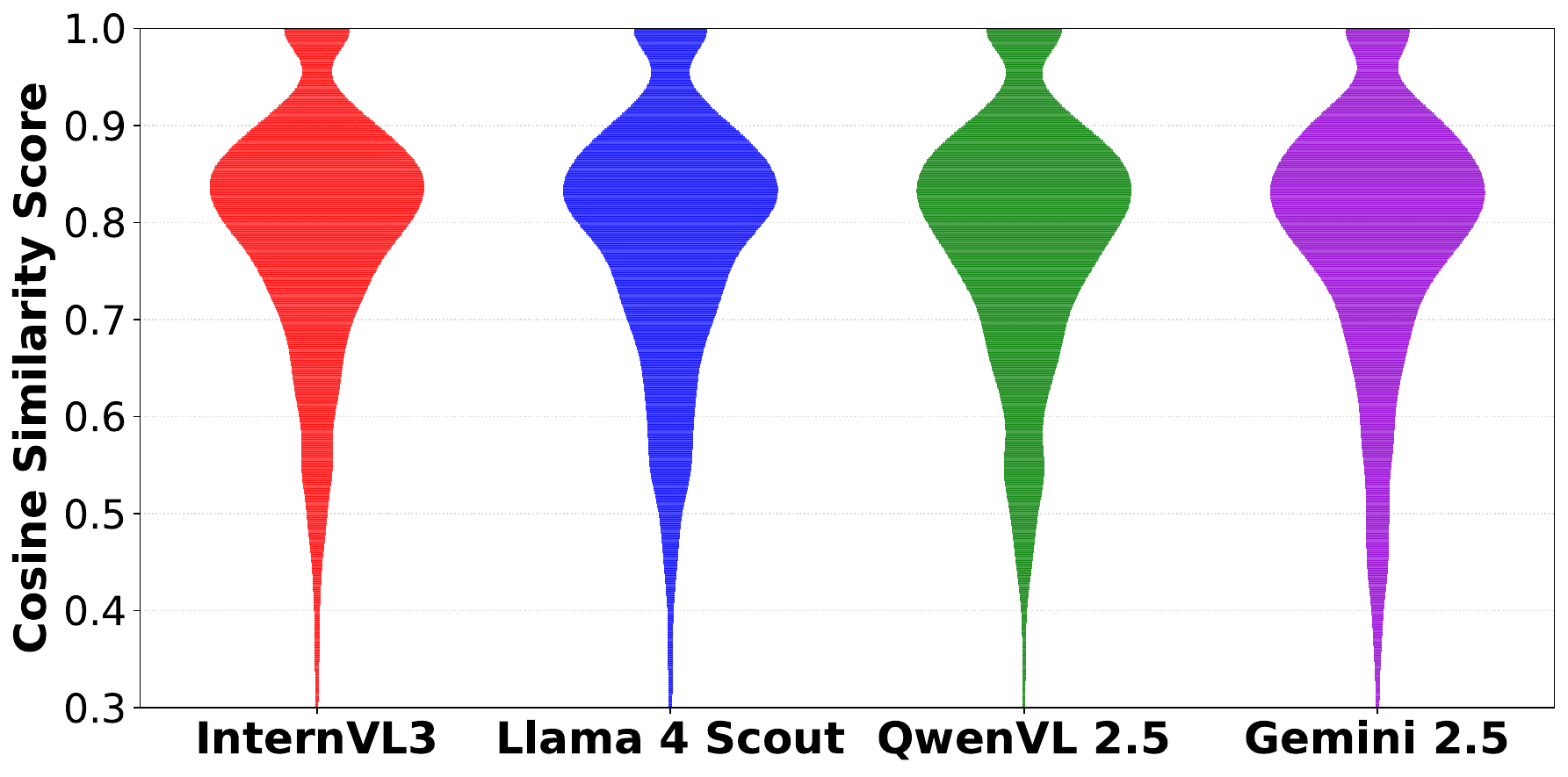}
    \caption{Density distribution of the cosine similarity scores between the concept word and the word predicted by the VLMs.}
    \label{fig:vlm_prediction}
\end{figure}

To further investigate the ability of VLMs to infer the intended concept from a metonymic image, we prompt each model with the image and ask it to generate the concept word it evokes. We then compute the cosine similarity between the BERT embedding of the generated word and that of the ground-truth concept word. This similarity score serves as a proxy for how closely the model’s output aligns with the intended metonymic meaning conveyed by the image. Figure~\ref{fig:vlm_prediction} presents the density distribution of cosine similarity scores across all evaluated models. We observe a consistent pattern: most scores fall within the 0.75 to 0.9 range. This high similarity suggests a key takeaway---VLMs can infer semantically similar terms, but it does not mean they have true human-level understanding, their performance is well below human on our ViMET dataset. It also reflects that our generated images are semantically close to the concept.

\section{CLIP Score of Metonymic Images}

\begin{figure}[h]
    \centering
    \includegraphics[width=0.95\linewidth]{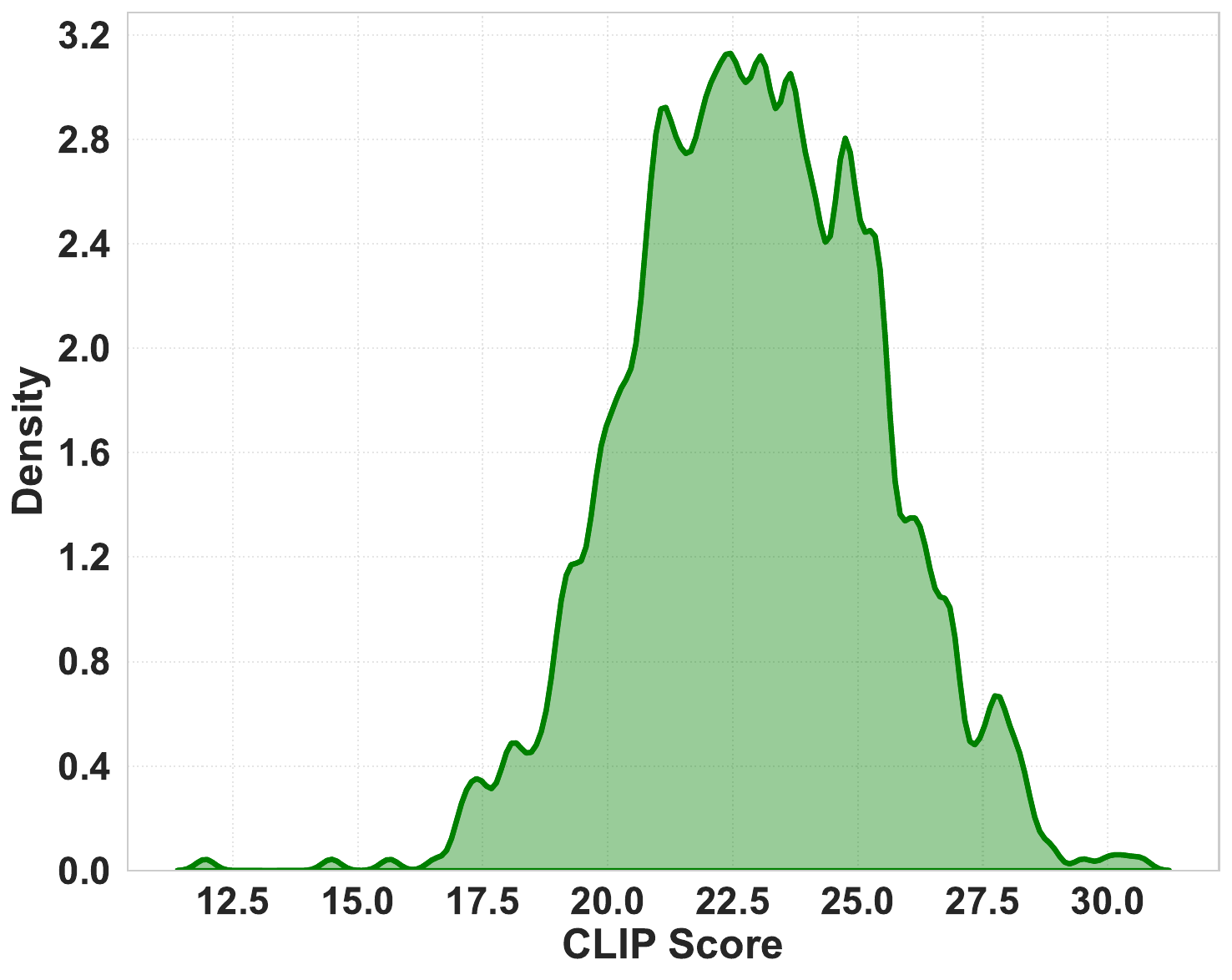}
    \caption{CLIP similarity score for the ViMET dataset.}
    \label{fig:clip_score}
\end{figure}

To quantify the semantic alignment between the generated metonymic images and their corresponding concept words, we compute the CLIP similarity score~\citet{clip} between each image and its associated concept word. Figure~\ref{fig:clip_score} presents a density plot of the normalized similarity scores across the ViMET dataset. The distribution reveals that most images achieve moderate CLIP similarity, with scores ranging between 20.0 and 25.0 and peaking around 22.5. This trend suggests that while the images are not literal representations (which typically produce higher CLIP scores), they still maintain a meaningful conceptual link to the target word—consistent with the nature of metonymic association. The absence of extremely low scores indicates that the images are not semantically unrelated, while the lack of high similarity scores near 30.0 confirms that they do not depict the concept directly. These findings further validate the metonymic quality of our dataset from a vision-language embedding perspective.

\section{Naturalistic vs Stylistic Images}

To test how visual metonymy operates—whether it depends more on the content side of the image or on the cognitive interpretation by the viewer—we generated two variants of each concept: a naturalistic version and a stylistic version. To quantify the relationship between the naturalistic and stylistic variants of our generated images, we employed CLIP to compute image–image similarity. Specifically, for each concept, we paired the corresponding images from the naturalistic and stylistic sets and extracted image embeddings using the CLIP ViT-B/32 encoder. We normalized these embeddings and computed cosine similarity scores for each pair, providing a measure of how closely the stylistic images preserved the semantic content of their naturalistic counterparts. 

Averaging across all matched pairs yielded an overall similarity score of \textbf{0.70}, indicating that significant compositional differences exist. However, while the two image sets are not identical and contain notable differences, both were able to successfully evoke the intended concept. As discussed in Section~\ref{discussion}, this finding underscores that visual metonymy operates less as a surface-level resemblance and more as a conceptual phenomenon, where meaning arises through the viewer’s interpretation of the image.

\section{Examples of Cultural Association}
\label{appendix_cultural_association}

\begin{figure*}
    \centering
    \includegraphics[width=0.98\linewidth]{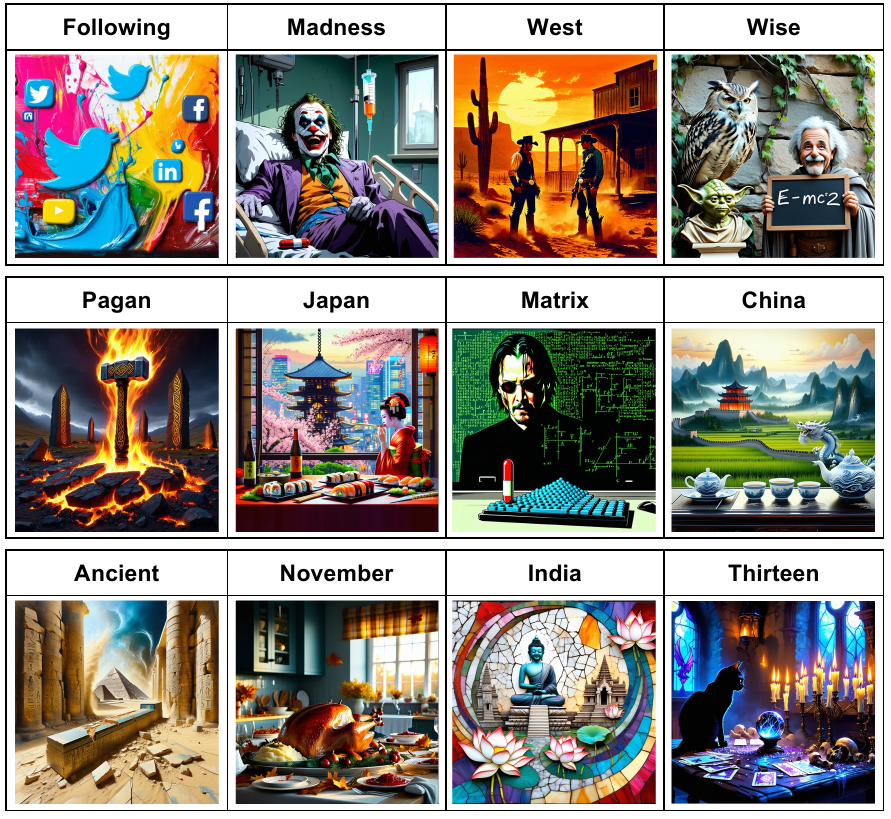}
    \caption{Example of visual metonymy images where the concept is connected to the image through cultural association.}
    \label{fig:cultural_example}
\end{figure*}

Figure~\ref{fig:cultural_example} shows examples where the concept is evoked from the image through cultural association. 

These associations draw on a variety of cultural domains: i) Societal culture, such as social media icons representing the concept \textit{following}; ii) National or country-specific culture, like turkey for \textit{November} (in the context of Thanksgiving), or sushi, cherry blossoms, and kimonos evoking \textit{Japan}; iii) Pop culture, such as gunslingers and desert landscapes for \textit{West}, the Joker for \textit{Madness}, or black cats and haunted objects for \textit{Thirteen} (linked to Friday the 13th).

In some cases, multiple cultural domains converge—for example, Yoda from Star Wars (pop culture) combined with a Wise Owl (societal culture) jointly evoke the concept \textit{Wise}.

\section{Examples of Contextual Association}
\label{appendix_contextual_association}

\begin{figure*}
    \centering
    \includegraphics[width=0.98\linewidth]{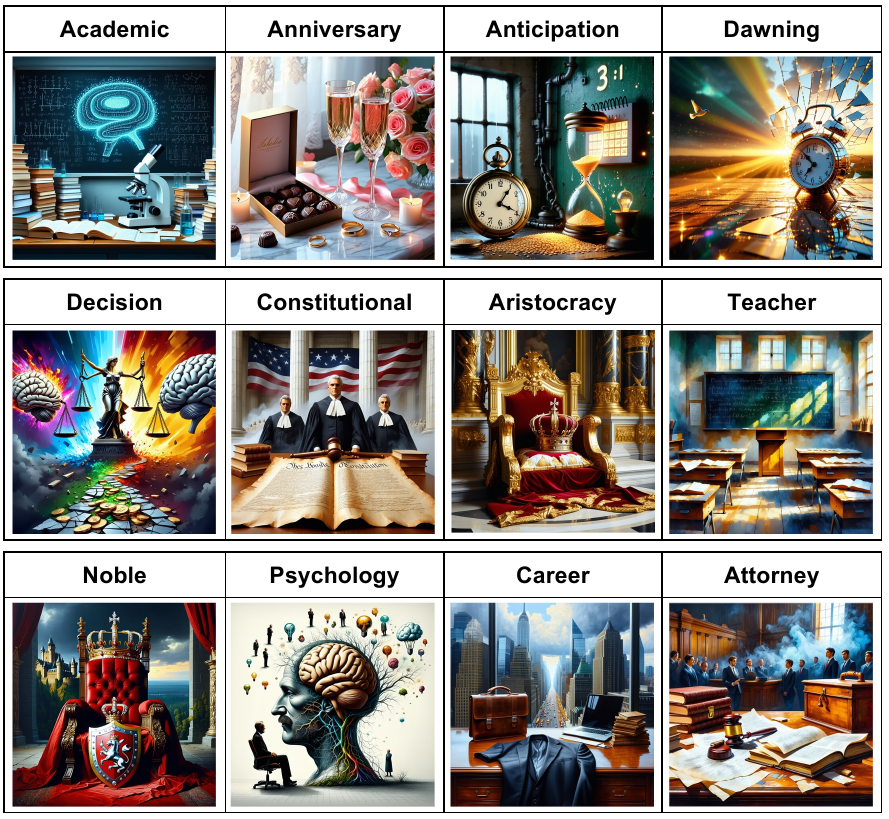}
    \caption{Example of visual metonymy images where the concept is connected to the image through contextual association.}
    \label{fig:contextual_example}
\end{figure*}

Figure~\ref{fig:contextual_example} shows examples where the concept is evoked from the image through contextual association.

As discussed in Section~\ref{conceptual_association_analysis}, these cases are harder to infer given the open nature of contextual association. The image refers to the concept, but do not point to it individually most of the time. For example, the image for the concept \textit{Academic} can also be mapped to concepts like \textit{Education} or \textit{Experimentation}. Similarly, the image for \textit{Aristocracy} can be identified with \textit{Monarchy} or \textit{Royalty}. This makes contextual associations harder to infer given the open nature of the association.

\section{Examples of Symbolic Association}
\label{appendix_symbolic_association}

\begin{figure*}
    \centering
    \includegraphics[width=0.98\linewidth]{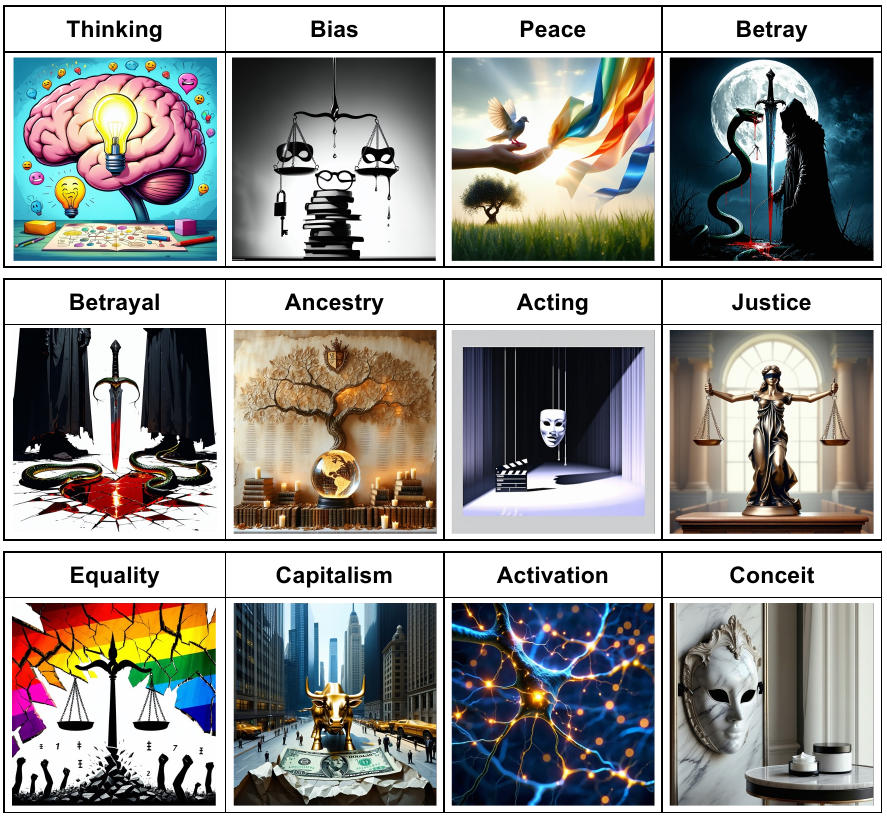}
    \caption{Example of visual metonymy images where the concept is connected to the image through symbolic association.}
    \label{fig:symbolic_example}
\end{figure*}

Figure~\ref{fig:symbolic_example} shows examples where the concept is evoked from the image through symbolic association.

These associations are typically the most direct to infer, provided the viewer is familiar with the meaning associated with the symbols. For instance, the statue of a bull accompanied by money invokes the idea of \textit{Capitalism} by referencing Wall Street iconography. The image for \textit{Activation} leverages the recognizable visual of firing neurons to symbolically convey the concept. Similarly, \textit{Justice} is represented by the classical symbol of a blindfolded woman holding balanced scales, while the concept of \textit{Bias} is expressed through the same scales—tilted unequally—to indicate imbalance.

\section{Examples Comparing Our Pipeline vs. General Method}
\label{direct_comparison_appendix}
\begin{figure*}
    \centering
    \includegraphics[width=0.98\linewidth]{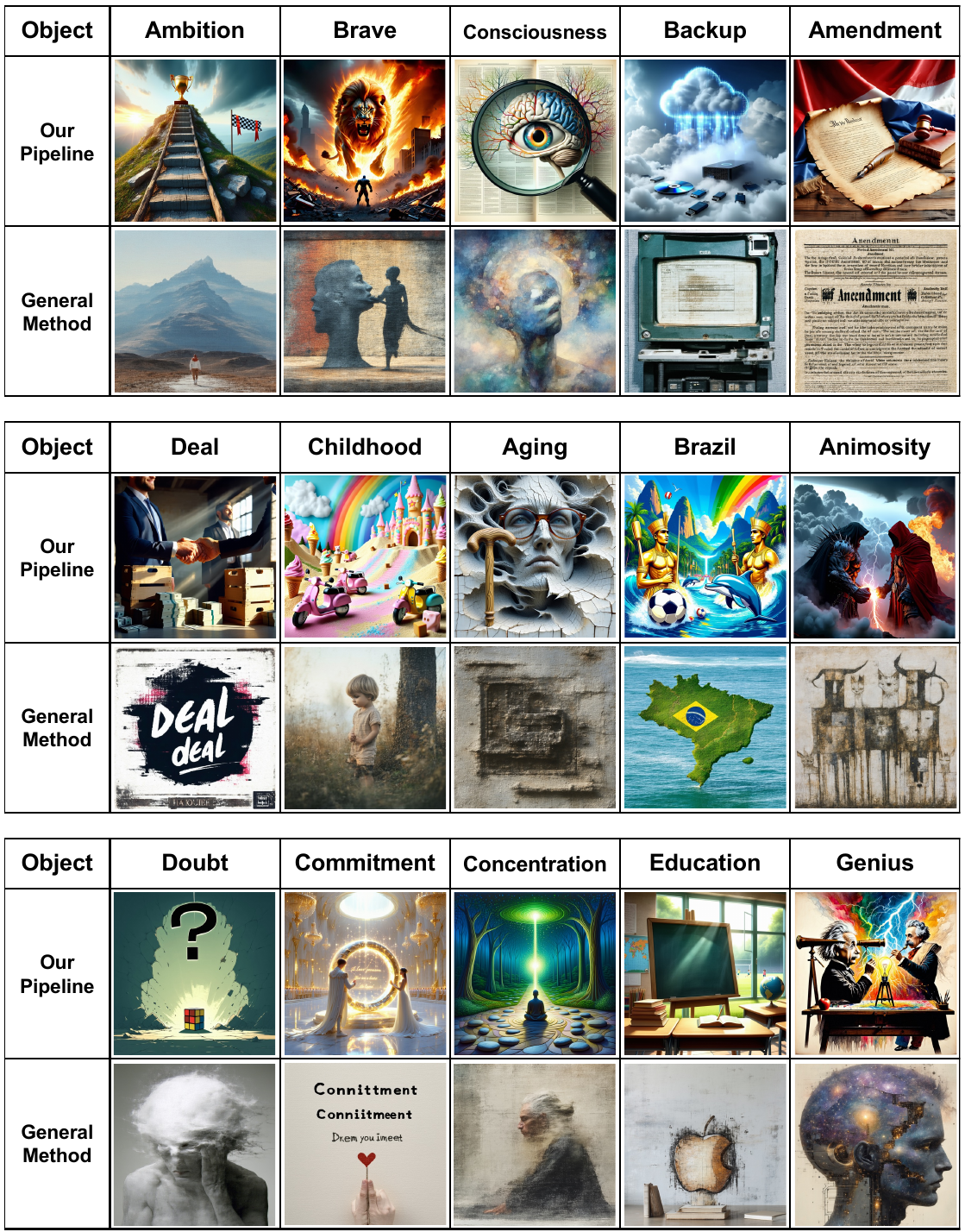}
    \caption{Example comparing images generated by our pipeline with images generating using a basic prompt.}
    \label{fig:appendix_comparison}
\end{figure*}

Figure~\ref{fig:appendix_comparison} shows examples comparing images generated by our pipeline with images generated using a basic prompt.

\section{Prompts Used in Our Pipeline} \label{app_prompts_used}
Table~\ref{prompts} shows the prompts used in our image generation pipeline.

\section{Example Output}
Table~\ref{tab:visual_description_example} provides examples of the representamens and visual descriptions generated by Llama using our pipeline.

\section{Annotation Guidelines}
\label{annotation_guidelines}
Figure~\ref{fig:annotation_guidelines_table} provides the annotation guidelines provided to the annotators. For the annotation task, we recruited students and colleagues. Their participation was completely voluntary.

\begingroup
\renewcommand{\arraystretch}{1}
\onecolumn
{\small
\begin{longtable}{|p{0.1\linewidth}|p{0.8\linewidth}|}
\caption{Prompts used in our image generation pipeline.}\label{prompts}\\

\hline
\bf Task & \bf Prompt \\
\hline
\endfirsthead

\multicolumn{2}{l}{\textit{(Continued from previous page)}}\\
\hline
\bf Task & \bf Prompt \\
\hline
\endhead

\hline
\multicolumn{2}{r}{\textit{(Continued on next page)}}\\

\endfoot

\hline
\endlastfoot

& \\
Generating Representamens & 

You are an expert in semiotics and conceptual associations. In Peirce’s semiotic triad, an object (concept) is linked to its representamen (representation or sign) through a commonly understood interpretation. Your task is to generate a list of representamen (signs) that people commonly associate with a given object.
Given an object, list 5 representamens (associations) that best reflect common cultural, contextual or perceptual links to the concept. The responses should be diverse, covering physical objects only that can be represented by an image, related to the object. Focus on common interpretations that most people would recognize. Give your answer in the following format:

Object: Vacation -> Representamen: Beach, mountains, airplane, suitcase, sun, resort

Object: Technology -> Representamen: Laptop, smartphone, Neural network diagram, Formula written on board, robots, circuits

Object: Paris -> Representamen: Eiffel Tower, Baguette, Notre Dame Cathedral, Parisian Architecture buildings, Bastille

Object: University -> Representamen: Books, classroom, black board, Campus, graduation cap

Object: Policeman -> Representamen: Police Uniform, Police Badge, Gun, Police car, Handcuffs

Object: \textit{{word}} -> Representamen:

\\

\hline

& \\

Generating Visual Prompt (Naturalistic Images)

&
       
You are an expert in visual metonymy and scene description. Given a set of objects that are associated with a broader concept, your task is to craft a short, vivid and immersive description of an image where these objects are the primary focus. The description should be ideally less than 70 words and should depict a scene that could be captured in a photograph or artwork. Use the following guidelines while generating the description of the image:

1. Do not describe the entity that these objects represent (e.g., if the objects relate to a doctor, do not mention the doctor—focus only on the objects).

2. The objects should be the focal point of your description, arranged in a way that clearly suggests the concept they are associated with.

3. Use rich sensory details (sight, lighting, textures, mood) to enhance realism, ensuring the scene conveys a strong metonymic connection to the broader concept.

4. Avoid using the concept word in your description.

Format your output in the following way:

Objects: beach, mountains, airplane, suitcase || Concept Word: Vacation || Output: An open suitcase sits on a wooden deck, its contents spilling slightly—a pair of sunglasses, a folded map, and a sand-dusted postcard. In the distance, a sleek airplane ascends into a pastel sky, framed by the silhouette of towering mountains. A gust of wind carries the scent of salt and pine, as if blending two destinations into one moment of anticipation.

Objects: soldiers, battlefield, tanks, explosions || Concept Word: War || Output: A vast battlefield stretches into the distance, covered in thick smoke and the echoes of distant gunfire. Soldiers in full combat gear move cautiously across the rugged terrain, their silhouettes outlined against the dim glow of burning wreckage. Tanks roll forward, their heavy treads leaving deep imprints in the mud. In the sky, an explosion erupts, sending a fiery plume of smoke into the air as debris scatters across the ground, capturing the chaos and intensity of war.

Objects: bird in flight, open chains, raised fists || Concept Word: Freedom || Output: Under a golden sunset, a bird soars high above the landscape, its wings spread wide in an effortless glide. Below, on a rocky cliff, an old pair of chains lies broken and open, their shattered links gleaming in the fading light. In the distance, a group of people stands together, their raised fists symbolizing resilience and liberation, silhouetted against the vast sky as the wind carries the echoes of their triumph.

Objects: \textit{rep\_input} || Concept Word: \textit{goal} || Output:
\\

\hline


Generating Visual Prompt (Stylistic Images)

&
       
You are an expert in visual metonymy and abstract scene description. Given a set of objects that are associated with a broader concept, your task is to craft a short, vivid and immersive abstract artistic description of an image where these objects are the primary focus. The description should be under 60 words and depict a scene rendered in the style of abstract art, such as Cubism, Futurism, or Surrealism, as if created by an artist like Picasso or Kandinsky. Follow these guidelines:

1. Do not name or describe the broader concept that these objects represent (e.g., if the objects relate to a doctor, do not mention the doctor—focus only on the objects).

2. The objects should be the focal point of your description. Arrange them in visually meaningful way that clearly suggests the concept they are associated with.

3. Use rich artistic details such as shape deformation, color dissonance, geometric rearrangement, fractured perspective, and expressive visual rhythm characteristics of the chosen abstract style.

4. Avoid abstract or symbolic references. The scene should feel an intentional abstract painting, not a figurative or symbolic description.

5. Do not use the provided concept word anywhere in the description.
Format your output in the following way:

Objects: beach, mountains, airplane, suitcase || Concept Word: Vacation || Output: Jagged triangles form a suitcase, its lid split open at impossible angles. Waves ripple in concentric cubes, colliding with faceted mountains stacked like shards of glass. An airplane arcs overhead—its body fractured into swirling spirals and razor-thin planes. Warm ochres clash with piercing teals, the entire canvas pulsing with disjointed motion and restless anticipation.

Objects: soldiers, battlefield, tanks, explosions || Concept Word: War || Output: Angular figures of soldiers twist in fragmented poses, limbs jutting through fractured planes of crimson and rust. Tanks are rendered as overlapping gears and steel blocks, scattered across a jagged terrain of burnt ochre. Explosions spiral outward in chaotic bursts of black and orange, the scene vibrating with sharp lines, shattered symmetry, and clashing momentum.

Objects: bird in flight, open chains, raised fists || Concept Word: Freedom || Output: The bird fractures mid-flight, wings stretched across intersecting arcs of turquoise and gold. Chains unravel in broken loops, spiraling outward like cracked halos. Fists rise in layered silhouettes—blocky, angular, and outstretched—each outlined in bold charcoal strokes. The canvas thrums with asymmetry, jagged rhythm, and defiant upward motion, captured in a swirl of raw, kinetic geometry.

Objects: \textit{rep\_input} || Concept Word: \textit{goal} || Output:

\\

\hline

\end{longtable}
}
\endgroup

\begingroup
\renewcommand\baselinestretch{0.9}

\NiceMatrixOptions{
  custom-line = {
    command = dashedmidrule,
    tikz = { dashed },
    total-width = \pgflinewidth + \aboverulesep + \belowrulesep,
  }
}

\begingroup
\renewcommand{\arraystretch}{2}
\onecolumn
{\small
\begin{longtable}{|p{0.075\linewidth}|p{0.15\linewidth}|p{0.3375\linewidth}|p{0.3375\linewidth}|}
\caption{Examples of representamens and visual descriptions generated by Llama using our framework.}\label{tab:visual_description_example}\\

\hline
\bf Concept Word & \bf Generated Representamens & \bf Generated Naturalistic Description & \bf Generated Stylistic Description \\
\hline
\endfirsthead

\multicolumn{4}{l}{\textit{(Continued from previous page)}}\\
\hline
\bf Concept Word & \bf Generated Representamens & \bf Generated Naturalistic Description & \bf Generated Stylistic Description \\
\hline
\endhead

\hline
\multicolumn{4}{r}{\textit{(Continued on next page)}}\\

\endfoot

\hline
\endlastfoot


Deal
&
1. Contract signed

2. Handshake

3. Exchange of 

goods

4. Currency

5. Transaction 

Completed
&
Two suited individuals stand opposite one another, clasping hands firmly as they exchange papers as contract now sealed between them. Nearby, several crates sit stacked neatly alongside piles of money, testament to the successful transaction just completed. Sunlight streams through nearby windows, illuminating the satisfied expressions on both men's faces.
&
Fractured sheets of paper curl and twist around interlocked hands, which cradle a pile of shimmering coins. Goods swirl and fragment into spiraled patterns, as money signs dance across a shattered backdrop of blues and reds. The entire tableau throbs with vibrant energy, rendered in the bold, expressive style of Expressionist abstraction.
\\

\hline


Training
&
1. Workout clothes

2. Dumbbells

3. Exercise mat

4. Gym equipment

5. Timer clock
&
A timer ticks down on the wall above a row of neatly stacked dumbbells. On the floor below sits a folded exercise mat, next to it a pile of clean workout gear. The room smells faintly of rubber and sweat, the space dim except for a few bright spotlights shining onto empty treadmills and weight machines. Somewhere nearby someone exhales loudly as a steady rhythm flows by.
&
In this chaotic gym room, all kinds of fitness tools fly around wildly - colorful dumbbells swirl above a tangled exercise mat; a timer clock ticks frantically amongst scattered weights; even workout clothes seem alive as they twist themselves into knots! All elements come together seamlessly to create one mesmerizing piece of modern art.
\\
\hline

Waiting
&
1. Clock ticking

2. Hourglass

3. Queue line

4. Traffic lights

5. Telephone 

ringing
&
An analog clock ticks loudly in an otherwise silent room, its second hand jerking forward one click at a time. On the wall behind it, a printed schedule lists times and destinations in neat columns. Beneath those rows, people sit slumped over in stiff chairs, heads drooping onto folded arms, eyes closed or staring blankly ahead. The air feels still and oppressive, weighed down by expectation and delay.
&
Tick-tock...The clock's hands move steadily, marking time as people shuffle forward in the queue. The hourglass slowly empties, grains cascading downward. Outside, cars sit idling at traffic lights, engines rumbling. Inside, phones ring unanswered, echoing through empty corridors. All around us, there is stillness - and yet everything seems to be moving towards some unseen finish line.
\\

\hline

Journey
&
1. Pathway

2. Footprints

3. Travelers bag

4. Passport

5. Globe
&
Gravel crunches softly underneath heavy boots as a solitary figure makes their way down a winding dirt path. Atop their shoulder sits a well-worn backpack, emblazoned with patches from countless adventures past. A weather-beaten passport peeks out from one pocket, while the other cradles a tattered guidebook to destinations unknown. Before them, an endless expanse of road beckons.
&
Alongside twisting pathways etched by the footprints of wanderers past, a worn traveler's pack rests beside a tattered passport, its pages bearing stamps from every corner of the globe. The world itself unfolds as a vast landscape of continents and oceans rendered in overlapping planes of vibrant colors and fractured perspectives, evoking the spirit of adventurous discovery through Cubist abstractions.
\\
\hline


Commit-

ment
&
1. Ring

2. Vows

3. Promise

4. Contract

5. Signature
&
Two hands clasped together tightly around a silver ring, fingers interlaced between rough ridges engraved along its band. Nearby sits a stack of papers covered with neat signatures and bold headings, promises made, vows exchanged. A single pen rests atop it all, ink still wet upon the page.
&
Intersecting rings twist into fractal spirals, their edges dissolving into shimmering planes of metallic blue. Vowels form jagged shapes amidst a torrent of angular lines, while promises crackle like bolts of lightning across the canvas. Contracts warp and bend into impossible geometries, signatures reduced to stark calligraphic gestures. The scene pulses with electric intensity.
 \\

\hline


Morality
&
1. Cross

2. Bible

3. scales of 

justice

4. Heart

5. Dove

6. Prayer hands

&
A jagged cross slices through shimmering scales, its edges blurring into fractured planes of black and white. In the background, a dove emerges from billowing clouds, its feathers transformed into twisting tendrils of light. A book shatters into shattered fragments, spilling words and images onto the canvas. Through this fractured landscape runs a pulsing red line - a heart beating with passion and conviction.
&
Against a backdrop of wispy clouds and deep indigo sky sits a simple wooden cross. Beside it, an ornate gold-scaled heart gleams softly; above both floats a dove with outstretched wings. At the base lies an ancient tome - leather-bound and embossed with intricate designs. Its pages flip slowly in the wind, revealing passages etched in fading ink.
\\

\hline


Conscious-

ness
&
1. Eye symbol

2. Mind map

3. Human brain

model

4. Brain activity

graph

5. Thought bubble

icon

& 
An intricate mindmap unfurls beneath a magnifying glass lens, its branches tracing pathways of thought through a vast network of ideas. Nearby, a human brain sits nestled within a transparent case, its folds illuminated by flickering waves of color representing neural activity. Above both lies a single eye, its iris dilated and unblinking, peering intently at this inner landscape of awareness.
&
A spiraling mindmap fractures into interlocking neurons, where a glowing thought bubble hovers above a vibrant brain scan. An eye symbol shimmers at the center, surrounded by fractured fragments of cognitive theory. The entire canvas pulses with shimmering neural patterns, painted in the psychedelic style of Surrealist abstraction.
\\

\hline

\end{longtable}
}
\endgroup

\begin{figure*}[h]
\centering
\noindent
\fbox{%
    \parbox{0.97\linewidth}{%
        \textbf{Annotation guidelines}

        \medskip
        Visual metonymy is a phenomenon where an image invokes a concept without explicitly showing it. \\
        Your task is binary classification. You will be given an image and the concept the image is supposed to represent. Your task is to identify if the image is a metonymic image, i.e., it successfully evokes the concept without explicitly showing or mentioning it. Here are some guidelines:

        \medskip
        1. The image should not show the concept explicitly, be it a physical entity (like any object or human), or an action (like teaching, running, studying, etc.). If it is shown explicitly, the image is not metonymic.

        \medskip
        2. The image should not also refer to the concept through words or text in the image. If the concept word is present in text, the image is not metonymic.
        
        \medskip
        3. If you see any potentially graphic or problematic image (in terms of violence, toxicity or harmful social content), mark them.
        
    }%
}
\caption{Annotation guidelines.}
\label{fig:annotation_guidelines_table}
\end{figure*}

\end{document}